\documentclass[smallextended]{svjour3}   

\usepackage[utf8]{inputenc}
\usepackage{todonotes}
\usepackage{hyperref}
\usepackage{url}
\usepackage{scrextend}
\usepackage[english]{babel}
\usepackage{subcaption}
\usepackage{tikz}
\usepackage{amsmath,amssymb}
\usepackage{graphicx}
\usepackage{caption}
\usepackage{adjustbox}
\usepackage[figuresright]{rotating}
\usepackage{natbib}
\usepackage{cite}

\smartqed

\newcommand{\hivecote}{\textsc{HIVE-COTE}}

\newcommand{\arima}{\textsc{ARIMA}}
\newcommand{\ourmethod}{TSER}

\graphicspath{figures}

\begin{document}
\title{Time Series Extrinsic Regression}
\subtitle{Predicting numeric values from time series data}
\author{Chang Wei Tan \and Christoph Bergmeir \and Fran\c{c}ois Petitjean \and Geoffrey I. Webb}

\institute{Chang Wei Tan \and Christoph Bergmeir \and Fran\c{c}ois Petitjean \and Geoffrey I. Webb \at Faculty of Information Technology\\
25 Exhibition Walk\\
Monash University, Melbourne\\
VIC 3800, Australia\\
\email{chang.tan@monash.edu,christoph.bergmeir@monash.edu,francois.petitjean@monash.edu,\\
geoff.webb@monash.edu}}

\maketitle

\begin{abstract}
This paper studies Time Series Extrinsic Regression (\ourmethod{}): a regression task of which the aim is to learn the relationship between a time series and a continuous scalar variable; 
a task closely related to time series classification (TSC), which aims to learn the relationship between a time series and a categorical class label.
This task generalizes  time series forecasting (TSF), relaxing the requirement that the value predicted be a future value of the input series or primarily depend on more recent values. 

In this paper, we motivate and study this task, and benchmark existing solutions and adaptations of TSC algorithms on a novel archive of 19 \ourmethod{} datasets which we have assembled. 
Our results show that the state-of-the-art TSC algorithm Rocket, when adapted for regression, achieves the highest overall accuracy compared to adaptations of other TSC algorithms and state-of-the-art machine learning (ML) algorithms such as XGBoost, Random Forest and Support Vector Regression.
More importantly, we show that much research is needed in this field to improve the accuracy of ML models. 
We also find evidence that further research has excellent prospects of improving upon these straightforward baselines. 
\end{abstract}

\section{Introduction}
In the past decade, there has been an increasing interest in time series analysis research, in particular time series classification (TSC) \citep{bagnall2017great, dau2019ucr,bagnall2015time,fawaz2019deep,dempster2019rocket,tan2020fastee} and time series forecasting (TSF) \citep{hyndman2018brief,makridakis1982accuracy,makridakis2000m3,makridakis2018m4,makridakis2020m4}. 
TSC is the task of predicting a discrete label that classifies the time series into some finite discrete categories \citep{bagnall2017great,dau2019ucr}.
On the other hand, TSF  aims to predict future values of a series based on recent or seasonal values.
It typically assumes that future values will more closely resemble recent values than those in the distant past.

Despite the thousands of papers published in both of these fields each year,
there has been little investigation of \emph{Time Series Extrinsic Regression} (\ourmethod{}), i.e. a task to predict numeric values that depend on the whole series, rather than depending more on recent than past values such as TSF.
The difference between TSC and \ourmethod{} is that TSC maps a time series to a finite set of discrete labels while \ourmethod{} predicts a continuous value from the time series.
For instance, TSC might classify an ECG signal as arrhythmia or normal, while \ourmethod{} could be used to predict a quantitative value such as the heart rate or respiratory rate of a patient 
\citep{pimentel2015probabilistic,pimentel2016toward,meredith2012photoplethysmographic,karlen2010capnobase} based on patterns in the ECG signal.
\ourmethod{} can be considered a special case of \emph{scalar-on-function regression (SoFR)} from the statistics community \citep{reiss2017methods, goldsmith2014estimator}, where the functional data is a time series.
SoFR considers a time series as functional data and builds statistical models to map functional data to a scalar response value.
In our case, we address the problem from a ML perspective, treating it as a \emph{regression} problem, taking time series data as the input and outputting a numeric value.

The term \emph{regression} has different meaning in different contexts.
In the broader machine learning context, \emph{regression} means predicting a continuous numerical value from a set of features \citep{segal2004machine,sammut2011encyclopedia}.
With respect to TSF, \emph{regression} usually means fitting the historical time series data with a regression model such as \arima{} \citep{box1970time} or Exponential Smoothing \citep{gardner1985exponential,hyndman2008forecasting,chatfield1978holt} models to forecast future values of the time series.
These TSF regression models typically heavily rely on recent or seasonal values, or sliding input windows of some form.  

In this work, we refer to the \ourmethod{} problem as a more general methodology of \emph{predicting a single continuous scalar value from a time series}.
We aim to predict values that can be either a continuation of the input time series or external to it and do not necessarily need to be a future value or depend on recent values.
In the case where predicting a future value of a series is of interest, then that becomes a TSF problem.
If predicting a finite discrete value is of interest, then that becomes a TSC problem.
We are interested in a more general task that lies in between the spectrum of these two tasks, which cannot be solved intuitively using models from these two tasks or SoFR.

\begin{figure*}
    \centering
    \resizebox{\columnwidth}{!}{
    	\input{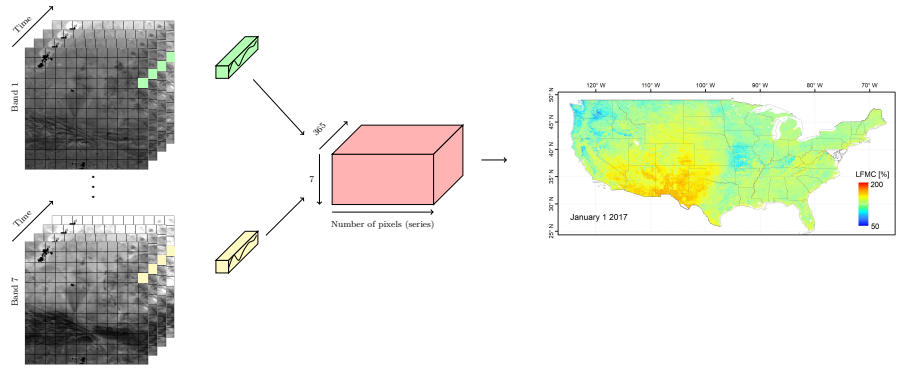}
    }	
    \caption{Prediction of live fuel moisture content (LFMC) using satellite images time series.
    }
    \label{fig:lfmc map}
\end{figure*}

For instance, we are interested in predicting the heart rate of a person from accelerometer data \citep{reiss2019deep,zhang2014troika}, predicting the crop yield or fuel load from satellite image series describing the evolution of the `colours' of the vegetation over the years; neither of which are discrete or future values.
Figure \ref{fig:lfmc map} shows the example of predicting live fuel moisture content (LFMC) of the United States using a series of satellite images where the value of LFMC is a continuous value in the range from 0 to 200$\%$. 
The input is the series of spectral values (i.e. time series of colour values) representing the state of a surface (or `pixel') over the last 12 months; the target is to infer the amount of moisture in the vegetation, i.e. the ratio between the weight of water in vegetation and the weight of the dry part of vegetation (information that is obtained by sampling vegetation in the field, weighing it and drying it to weigh it again). 
This is a very important variable, as the risk of fire increases very rapidly as soon as the LFMC goes below 80\% \citep{yebra2018fuel}, making it an invaluable variable for forest fire early warning systems. 
A very similar application is the one of predicting crop yield from these same series of spectral values, with great importance for food safety and agricultural planning \citep{pelletier2019temporal}. 

Typical regression algorithms do not work well when applied directly to such problems because they do not take into account the temporal aspect of the data. 
These algorithms also suffer from the curse of dimensionality, especially when the data is sampled with high sampling frequency and with a large number of channels.
TSC algorithms on the other hand were not designed for these continuous scalar outputs. In particular, they are predicated on the assumption that the output values are not ordered.
Hence, we need algorithms that are able to learn the relationship between time series data and the continuous scalar variable.
There has been some research in this area where the algorithms and features are specifically designed for the specific tasks \citep{reiss2019deep,zhang2014troika,zhang2015photoplethysmography,de2008field}.
Unfortunately, these algorithms do not generalise well to other problems.
For instance, those specific features created from photoplethysmogram (PPG) measurements \citep{zhang2014troika,reiss2019deep} for heart rate estimation cannot be used to predict crop yields and vice-versa. 

Therefore in this paper, we aim to motivate the research into developing more general \ourmethod{} algorithms.
We start by introducing the first \ourmethod{} benchmarking archive, which we have assembled and contains 19 datasets in various domains in \citep{tan2020monash}.
These datasets have varying number of dimensions, dimensions with unequal lengths and missing values.
They are used to benchmark some adaptations of classical regression and TSC algorithms as well as SoFR techniques. 
Our results show that simple variants of some state- of-the-art TSC algorithms outperform standard regression techniques (i.e. ones developed for tabular data) that do not take into account the underlying series nature of the data. 
More importantly, we show that most methods obtain similar accuracies and the top method --~Rocket~-- is actually not far in accuracy from algorithms that ignore the sequential information in the series data, XGBoost \citep{chen2016xgboost} and Random Forest \citep{breiman2001random}, which motivates the need for the development of a subfield of research. 

The rest of this paper is organised as follows.
In Section \ref{sec:background}, we introduce the problem that we aim to address and discuss the related work.
Then we describe some of the applications of \ourmethod{} with respect to the benchmark datasets we created in Section \ref{sec:datasets}.  
Section \ref{sec:algorithms} then describes how the classic regression and TSC algorithms can be adapted for \ourmethod{}.
After that, we evaluate these algorithms on the first \ourmethod{} benchmark datasets in Section~\ref{sec:results}.
Finally, in Section~\ref{sec:conclusion}, we summarise our contribution and give some direction for future work.

\section{Time Series Extrinsic Regression}
\label{sec:background}
\emph{Time Series Extrinsic Regression} (\ourmethod{}) is a regression task that learns the mapping from time series data to a scalar value.
It shares resemblance to other fields such as SoFR and time series regression, which has different meaning in different contexts.
In this section, we give a formal definition to \ourmethod{} as we employ it.
We will also try to clear any misunderstandings that the readers might have and introduce the task that we aim to address.
We first define a time series in Definition \ref{def:time series}. 

\begin{definition}
\label{def:time series}
A time series $S$ is an ordered collection of $L$ pairs of measurements and timestamps,
$S=\{(s_1,t_1),(s_2,t_2), ..., (s_L,t_L)\}$, 
where $s_i\in\mathbb{R}^D$ and $t_1$ to $t_L$ are the timestamps for some measurements $s_1$ to $s_L$. 
\end{definition}

\noindent
Note that the $D$-dimensional measurement $s_i$ measures the same phenomena with different instruments at the same time.
Time series data differs from static data in a way that the ordering of the data attribute in time series data is critical in finding the best discriminating features in time series data.

\emph{Classification} and \emph{Regression} are both supervised learning tasks that learn the relationship between a target variable and a set of features \citep{sammut2011encyclopedia}.  
The main difference between \emph{Classification} and \emph{Regression} is that  \emph{Classification} predicts a categorical value for a data instance that categorises the data into some finite categories, while \emph{Regression} predicts a continuous value.
\emph{Regression} tasks can become \emph{Classification} tasks when the predicted values are discretized into some finite labels for the data.
In this work, we only focus on \emph{Regression}.
For example, the simplest regression algorithm, linear regression, assumes a linear relationship between a set of predictors (features) and a target variable, and fits a straight line through all the predictors to generate a prediction for the target variable. 

Traditionally in ML, the features used for regression are static and have no relation to time.
For instance, we could predict house prices using features such as the number of bedrooms, crime rate, nitric oxides concentration (pollution level), accessibility to radial highways and weighted distances to employment centers \footnote{\url{https://www.kaggle.com/vikrishnan/boston-house-prices}}.
These features (predictors) do not depend on time and are less likely to change over time.
They are then used to train an ML algorithm such as a Random Forest \citep{breiman2001random}, XGBoost \citep{chen2016xgboost} or even linear regression to predict house price, the target variable that we are interested in. 
Different from the traditional regression problem, the regression problem that we tackle in this work, considers time series data as the features.
With respect to the house price prediction example, 
instead of using a single value for the number of rooms, crime rate or pollution level,
we use the time series of these features to predict house prices.
For example the daily crime rate or daily pollution level over the last one month.
A more concrete example of regression in our context is the prediction of heart rate which can only be achieved using time series data such as PPG and accelerometer data \citep{reiss2019deep,zhang2015photoplethysmography,zhang2014troika} that measures the pulse and movement of the subject within a certain period of time. 

\begin{figure}
    \centering
    \includegraphics[width=\columnwidth]{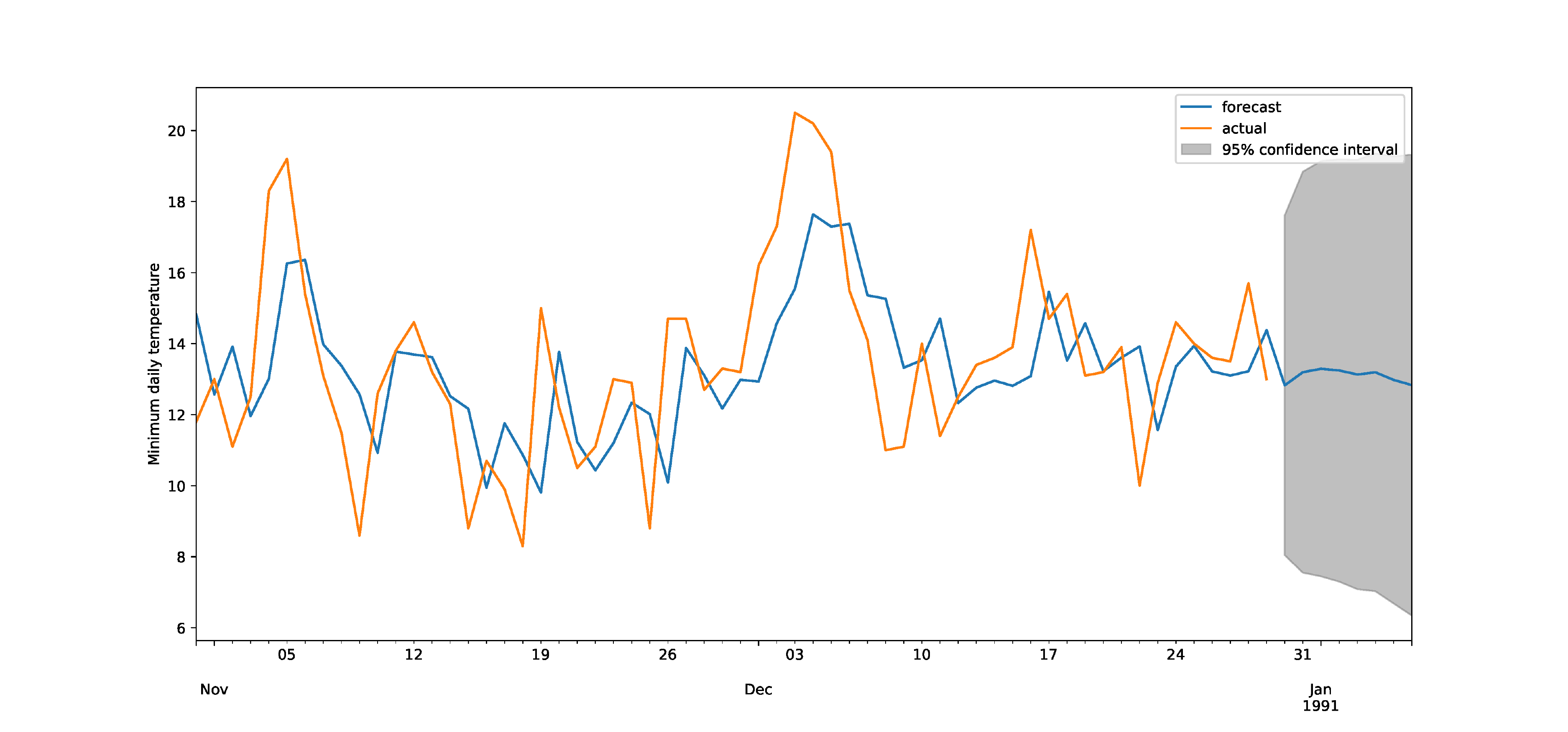}
    \caption{Example of an autoregression model of order 7, AR(7).}
    \label{fig:ar 7}
\end{figure}

A very large branch of time series analysis deals with TSF \citep{hyndman2018brief,hyndman2008forecasting,makridakis2018m4}, where \emph{regression} carries a slightly different meaning.
In TSF, \emph{regression} is used to fit autoregressive models on the historical time series which models the recent and/or seasonal values in the time series.
Figure \ref{fig:ar 7} shows an example of a linear autoregressive model of order 7, AR(7), i.e. the model uses the past 7 days minimum daily temperature to forecast the minimum daily temperature for the next day.
These models are then extrapolated to predict future values of the same time series. 
Going back to the example of predicting house prices, 
autoregressive models can be used to fit past house prices data and produce a good forecast for future house prices, as it is very likely that house price depends on the price in the previous months.
In our regression context, we can also build models to predict future house price using past house prices.
However, we aim at developing more general models that do not make the assumptions that frequently underlie forecasting models, such as that the most recent values are most indicative of future values.
In other words, we can see that forecasting models will not be useful in our regression example of predicting heart rate, as heart rate is not a future value of ECG, PPG and accelerometer signal and does not depend more on the final value of these data than on the initial ones. 

Rather, heart rate is a quantitative value of the signal that can be obtained through counting the number of peaks in the signal.
Formally, we define the task of \emph{Time Series Extrinsic Regression} in Definition \ref{def:tsr}.

\begin{definition}
\label{def:tsr}
A \emph{time series extrinsic regression model} is a function $\mathcal{T}\rightarrow \mathcal{R}$, where $\mathcal{T}$ is a class of time series. \emph{Time series extrinsic regression} seeks to learn a regression model from a dataset $\mathcal{D}=\{(t_1,r_1), \ldots, (t_n, r_n)\}$, where $t_i$ is a time series and $r_i$ is a continuous scalar value.
\end{definition}

\subsection{Related work}
\label{subsec:related work}
Time series data can be considered as functional data, where the measurements are a function of time \citep{goldsmith2014estimator}. 
Functional regression is a widely studied task in the statistics community 
\citep{reiss2017methods,goldsmith2014estimator}.
Functional regression models can be classified into three categories: (1) scalar responses with functional predictors (scalar-on-function regression); (2) functional responses with scalar predictors (function-on-scalar regression); and (3) functional responses with functional predictors (function-on-function regression) \citep{reiss2017methods}.
The task of mapping a time series to a scalar value, \ourmethod{}, is closely related to scalar-on-function regression (SoFR), a task that maps functional data (e.g., a time series) to a scalar response \citep{reiss2017methods,goldsmith2014estimator}.
SoFR typically works by first representing the time series data in its functional form. 
Then a basis function such as Functional Principal Components (FPC), B-spline, Fourier or Wavelet can be applied to smooth the data and reduce noise.
Finally a regression model is applied to the smoothed data to predict the scalar value.

Functional linear models (FLM) are the most common approach for SoFR, which extend the standard multiple linear regression model to functional data \citep{goldsmith2014estimator}.
Most work in the literature of SoFR focused on better estimating the weights that are applied to every timestep of the time series data \citep{goldsmith2014estimator}. 
The study of \citet{goldsmith2014estimator} shows that SoFR models have been applied to problems such as predicting annual rainfall from observed temperature and predicting fat content in meat from near-infrared spectrum. 
The study compares various SoFR models with its ensemble counterparts and non-functional models such as random forest and gradient boosting machines. 
The results concluded that ensembles of models work better than a single model.
More importantly, the results also show the limitation of FLMs where non-functional models such as random forest are robust and consistently outperform other FLMs on all the test datasets.
In addition, functional regression models usually require an in-depth understanding of the data on hand and experience, in order to apply the right basis function to fit the model.
For instance, Fourier basis functions will not work well on non-periodic signals.

While we have not been able to identify any prior work in the ML community specifically addressing the more general class of learning task that we call \emph{time series extrinsic regression}, there are a number of specialised techniques addressing specific cases. 
In addition to forecasting, one that has received considerable attention is heart rate (HR) estimation using photoplethysmogram (PPG) sensors \citep{reiss2019deep,zhang2014troika}. 
These methods rely on spectral analysis \citep{zhang2014troika,zhang2015photoplethysmography,salehizadeh2016novel,schack2017computationally} but they were not very accurate \citep{reiss2019deep}.
A convolutional neural network based approach that takes the signal in the frequency domain as input has been proposed  to improve the prediction accuracy \citep{reiss2019deep}.
This approach was shown to be significantly more accurate compared to the existing spectral methods.

Similar to heart rate estimation, respiratory rate (RR) estimation can also be achieved using PPG sensors \citep{pimentel2016toward,meredith2012photoplethysmographic,pimentel2015probabilistic}. 
Estimating RR is an important task because it is often the earliest sign of critical illness \citep{meredith2012photoplethysmographic}.
Existing methods fail to distinguish between periods of high and low quality data and were not able to generalise well to other datasets \citep{pimentel2016toward}.
Typically, estimation of RR from PPG is achieved by applying a moving window to the time series producing an estimate for RR per window \citep{pimentel2016toward} and consists of four key components, (a) extracting respiratory signals; (b) estimating respiratory rates; (c) fusing the estimates and (d) quality assessments \citep{pimentel2015probabilistic,pimentel2016toward}. 
A probabilistic approach was proposed  \citep{pimentel2015probabilistic}  using the Gaussian process regression framework to extract RR from the different sources of modulation in the PPG signal.
The authors then proposed another method  \citep{pimentel2016toward} by fitting multiple autoregressive models to the extracted respiratory signals. 
Their method was evaluated on two datasets, the Capnobase \citep{karlen2010capnobase} and the BIDMC dataset \citep{pimentel2016toward} (both can be found in \url{http://peterhcharlton.github.io/RRest/datasets.html}).
Although the results showed that their method achieved the best mean absolute error (MAE) on both datasets compared to other existing methods in RR estimation, it was only significantly different to one of the methods on the Capnobase dataset.
There were no significant difference on the BIDMC dataset.

Other than health monitoring, there are also similar works done for pollution monitoring, where the goal is to predict pollutant concentration using on-field sensors \citep{de2008field}.
\citet{de2008field} proposed a simple feed-forward network with 5 hidden layers, taking 7 sensor inputs to estimate benzene concentration in an Italian city.
The method, although simple, achieved very low MAE of $0.13\mu g/m^3$, but is not generalisable.

\subsection{\ourmethod{} applications and datasets}
\label{sec:datasets}
To support research \ourmethod{}, we created the first \ourmethod{ benchmarking archive, available online at \url{http://tseregression.org/}.
In this section, we describe the possible applications of \ourmethod{} and our first \ourmethod{} archive.
The current \ourmethod{} archive contains 19 time series datasets from 5 application areas, \emph{Health Monitoring}, \emph{Energy  Monitoring}, \emph{Environment Monitoring}, \emph{Sentiment Analysis} and \emph{Forecasting}.
The archive contains 8 datasets assembled from the UCI machine learning repository \citep{Dua:2019}, 3 from physionet.org, 1 from a signal processing competition \citep{zhang2014troika}, 1 from the Covid-19 database from the World Health Organisation, 1 from the Australian Bureau of Meteorology (BOM) and the rest are donations.
These datasets are unnormalised with varying number of dimensions, unequal length dimensions and missing values.
We briefly describe these datasets below and refer readers to \citep{tan2020monash} for a more detailed description.
Table~\ref{tab:datasets} outlines the properties of the datasets in the current \ourmethod{} archive. 

\begin{table}[]
	\centering
	\resizebox{\textwidth}{!}{
		\begin{tabular}{|l|l|c|c|c|c|c|c|}
			\hline
			Id & Dataset & Train size & Test size & Length & No of Dimension & Missing \\
			\hline
			\multicolumn{7}{|c|}{Energy Monitoring} \\
			\hline
			1 & AppliancesEnergy & 96 & 42 & 144 & 24 & No \\ 
			2 & HouseholdPowerConsumption1 & 746 & 694 & 1440 & 5 & Yes \\
			3 & HouseholdPowerConsumption2 & 746 & 694 & 1440 & 5 & Yes\\
			\hline 
			\multicolumn{7}{|c|}{Environment Monitoring} \\
			\hline
			4 & BenzeneConcentration & 3433 & 5445 & 240 & 9 & Yes \\
			5 & BeijingPM25Quality & 12432 & 5100 & 24 & 9 & Yes \\
			6 & BeijingPM10Quality & 12432 & 5100 & 24 & 9 & Yes \\
			7 & LiveFuelMoistureContent & 3493 & 1510 & 365 & 7 & No \\
			8 & FloodModeling1 & 471 & 202 & 266 & 1 & No \\
			9 & FloodModeling2 & 389 & 167 & 266 & 1 & No \\
			10 & FloodModeling3 & 429 & 184 & 266 & 1 & No \\ 
			11 & AustraliaRainfall & 112186 & 48081 & 24 & 3 & No \\
			\hline
			\multicolumn{7}{|c|}{Health Monitoring} \\
			\hline
			12 & PPGDalia* & 43215 & 21482 & 256,512 & 4 & No \\ 
			13 & IEEEPPG & 1768 & 1328 & 1000 & 5 & No \\
			14 & BIDMCRR & 5471 & 2399 & 4000 & 2 & No \\
			15 & BIDMCHR & 5550 & 2399 & 4000 & 2 & No \\
			16 & BIDMCSpO2 & 5550 & 2399 & 4000 & 2 & No \\
			\hline
			\multicolumn{7}{|c|}{Sentiment Analysis} \\
			\hline 
			17 & NewsHeadlineSentiment & 58213 & 24951 & 144 & 3 & No \\
			18 & NewsTitleSentiment & 58213 & 24951 & 144 & 3 & No \\
			\hline
			\multicolumn{7}{|c|}{Forecasting} \\
			\hline
			19 & Covid3Month & 140 & 61 & 84 & 1 & No \\
			
			\hline
		\end{tabular}
	}
	\caption{Time series datasets in the current \ourmethod{} archive. The ones marked with an asterisk (*) have different lengths from one dimension to another (but the length is the same for all instances in any single dimension).}
	\label{tab:datasets}
\end{table}

\subsubsection{Energy monitoring}
With advances in Smart City and Internet of Things applications, the task to monitor energy and power consumption has become more important than ever.
The ability to predict energy and power consumption accurately can save millions of dollars for a big company. 
Energy monitoring is typically done by collecting data such as temperature, humidity, rain, voltage and current readings from sensors attached all over a building. 
These data are collected in the form of time series and is mapped to the power consumption of the building.
For example, higher power consumption will be observed during winter months as more energy is required to heat up a building. 
The \textbf{AppliancesEnergy}, \textbf{HouseholdPowerConsumption1} and \textbf{HouseholdPowerConsumption2} are the three datasets in this archive targeting this application. 
Figure \ref{fig:household power} shows an example of time series data in the HouseholdPowerConsumption datasets. 

\begin{figure}[!t]
    \centering
    \includegraphics[width=0.8\columnwidth]{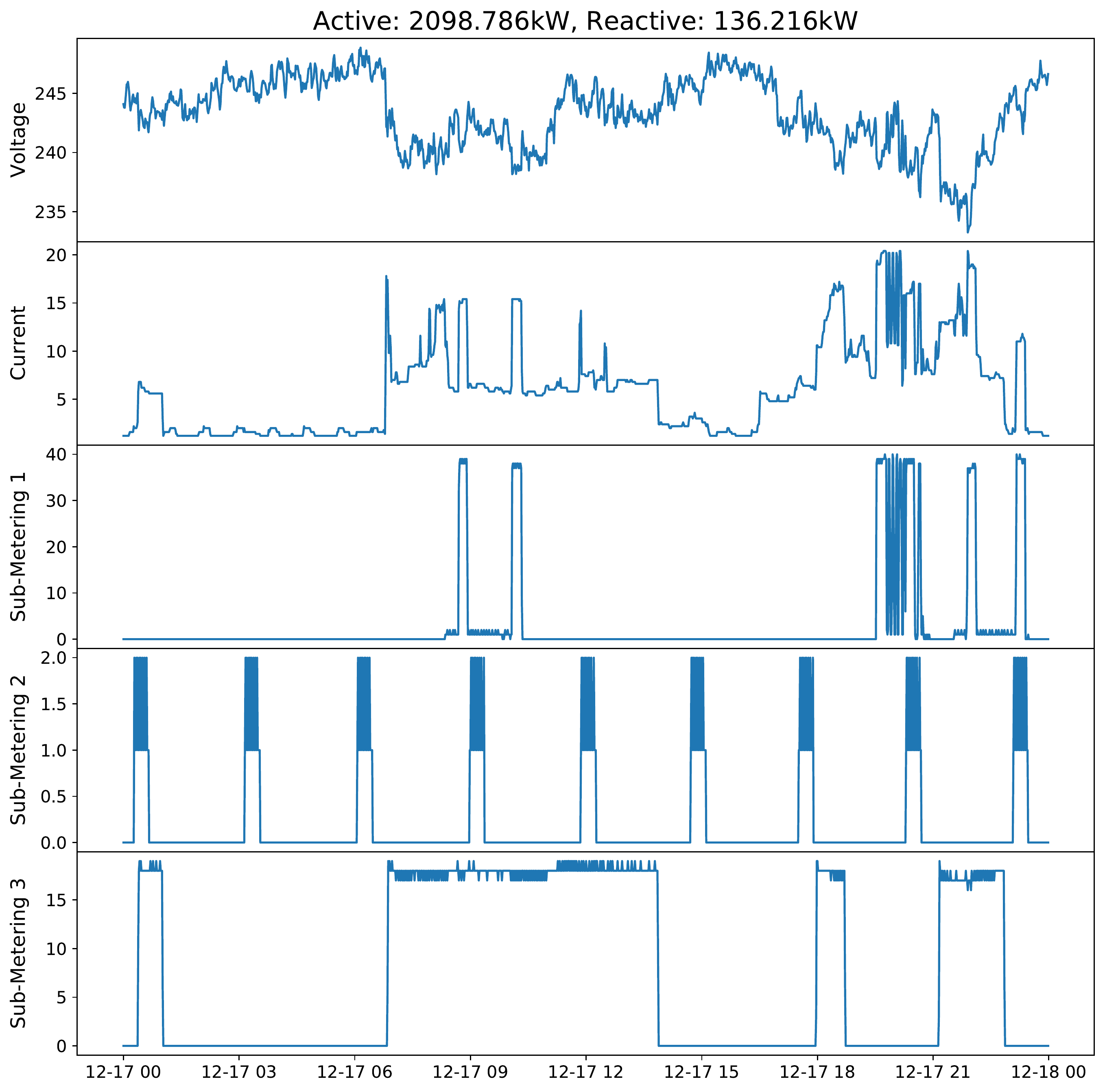}
    \caption{Examples of the daily voltage, current and sub-metering measurements in the HouseholdPowerConsumption dataset that is used to predict the total daily active and reactive power consumption in a house.}
    \label{fig:household power}
\end{figure}

\subsubsection{Environment monitoring}
In the context of climate change, environment monitoring has become more important than ever.
Environment monitoring is the task of predicting anything related to our environment such as pollution level, rainfall, crop yield and flood water level.
The three datasets \textbf{BenzeneConcentration}, \textbf{BeijingPM10Quality} and \textbf{BeijingPM25Quality} focus on predicting pollution level in a metropolitan city.
The \textbf{LiveFuelMoistureContent} is a dataset about predicting live fuel moisture content (moisture content in leaves) using series of satellite images, which we described in the introduction. 
Predicting the moisture content is very critical in bushfire prevention that could prevent the lost of thousands of lives and millions to billions of dollars.
Figure \ref{fig:lfmc} shows an example of the satellite image time series of a tree cover with 7 spectral bands in the LiveFuelMoistureContent dataset. 
The three \textbf{FloodModeling} datasets address prediction of the height of different riverbeds given a series of rainfall events. Here again, being able to predict the rise of water is critical to mitigate its risk. The relationship between rainfall and water height in different locations is non-linear, as it depends on topography, transpiration and rainfall dynamics. Here we assume that topography and land-cover (which drives transpiration) is not known and propose to model water height directly from rainfall time series. 
Finally, the \textbf{AustraliaRainfall} dataset contains the hourly temperature of various locations in Australia and the goal is to predict the total daily rainfall in those locations based on the hourly temperature. This is useful as temperature sensors are much cheaper and easy to maintain as compared to rain gauges. 

\begin{figure}[!t]
    \centering
    \includegraphics[width=0.8\columnwidth]{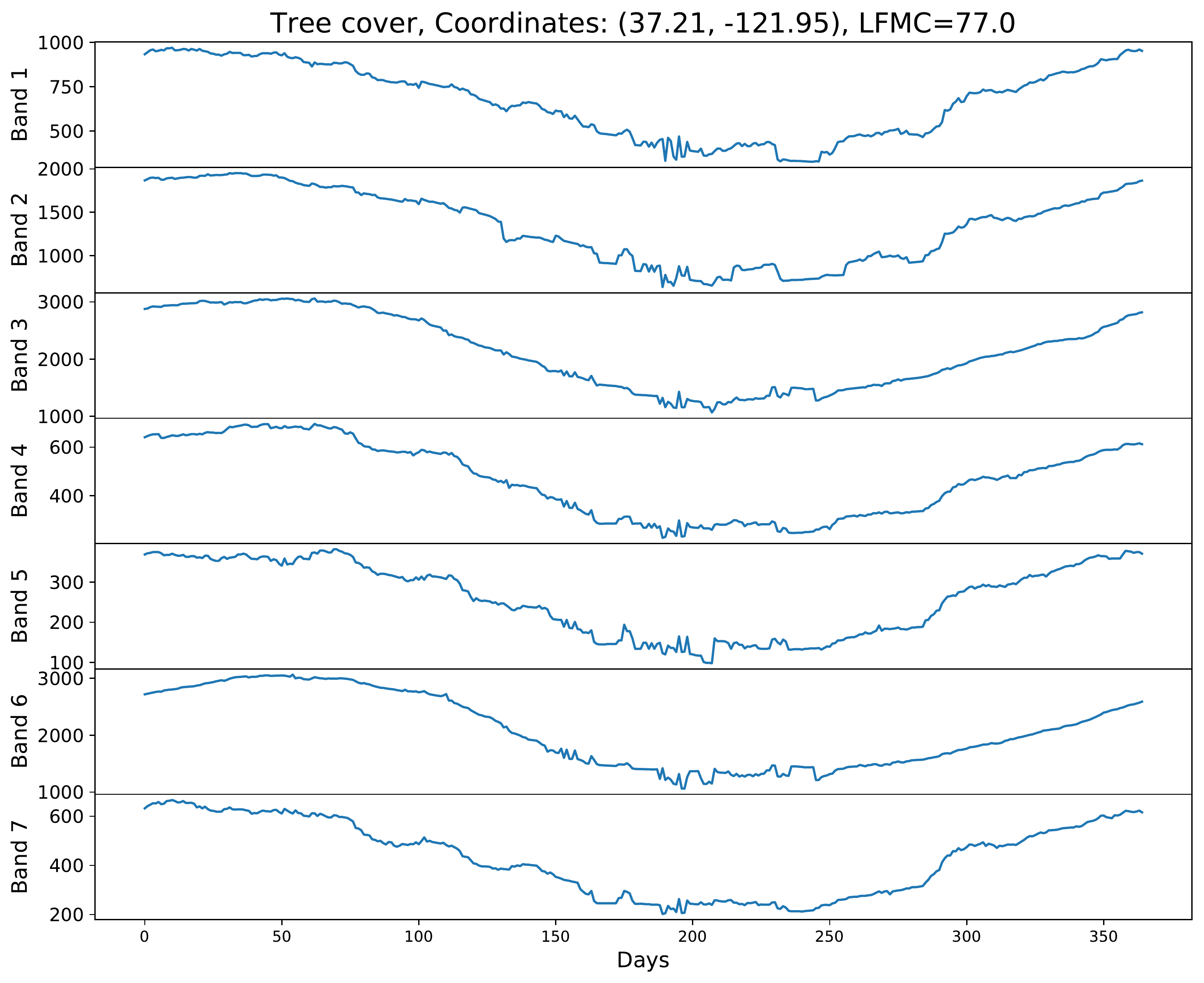}
    \caption{Example of LiveFuelMoistureContent time series with 7 spectral bands.}
    \label{fig:lfmc}
\end{figure}

\subsubsection{Health monitoring}
Health monitoring is the task of monitoring the health or vital signs of an individual.
The data typically comes from a wearable device that can be attached to the subject, such as a photoplethysmogram (PPG), electrocardiogram (ECG), electroencephalogram (EEG) or accelerometer. 
In this work, we focus on three tasks, estimating heart rate, respiratory rate and blood oxygen saturation level. 
The \textbf{PPGDalia}, \textbf{IEEEPPG} and \textbf{BIDMCHR} are datasets focusing on heart rate estimation. 
Figure \ref{fig:ppg_dalia} illustrates an example of the PPG and accelerometer signal from the PPGDalia dataset.  
\textbf{BIDMCRR} and \textbf{BIDMCSpO2} are both datasets on predicting respiratory rate and blood oxygen saturation level, respectively.

\begin{figure}[!t]
    \centering
    \includegraphics[width=0.65\columnwidth]{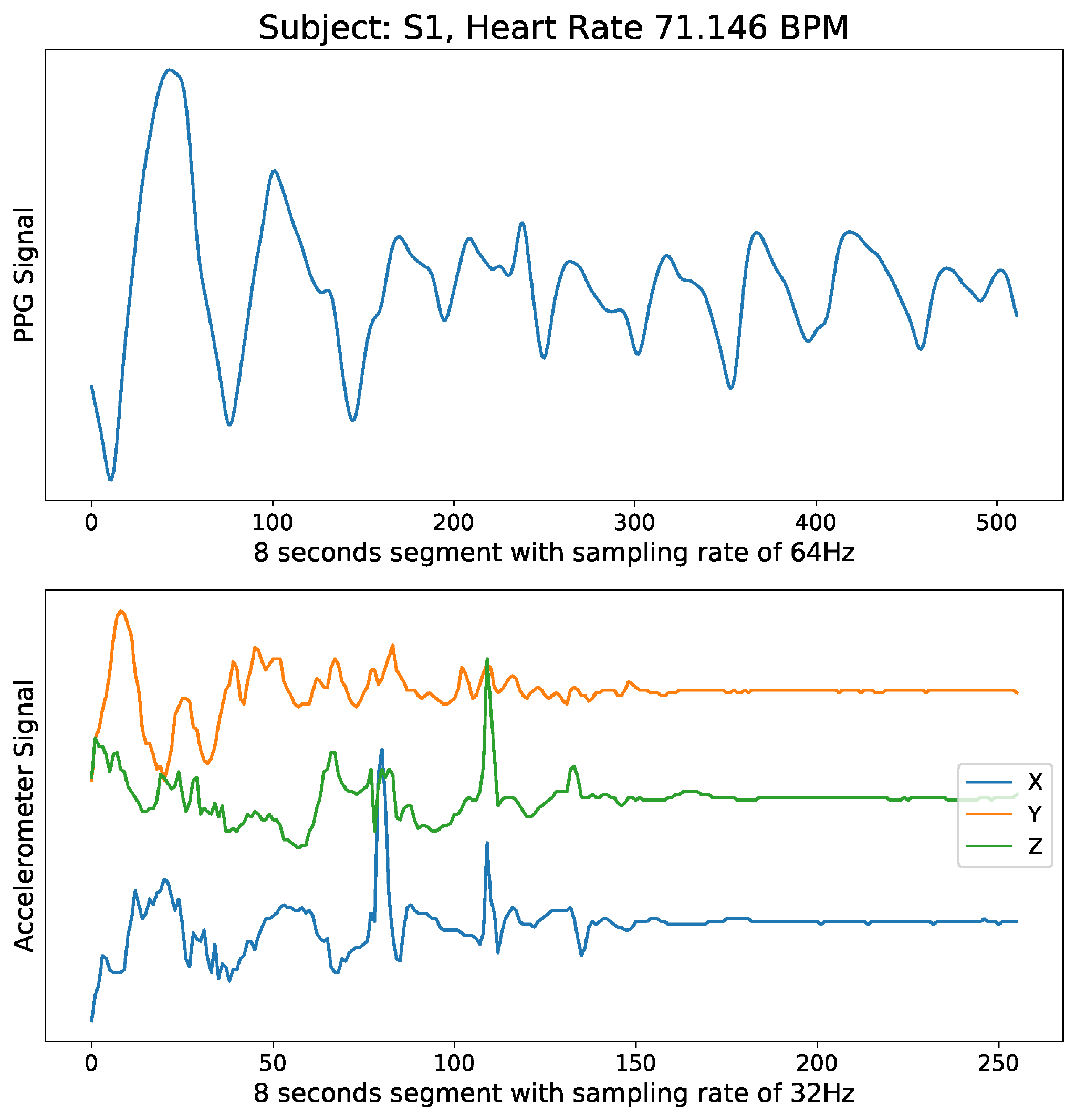}
    \caption{Example of time series in the PPGDalia dataset. The title shows the subject and the current heart rate in beats per minute (BPM).}
    \label{fig:ppg_dalia}
\end{figure}

\subsubsection{Sentiment analysis}
Sentiment analysis is the interpretation and classification of emotions (positive, negative or neutral) within some text using text analysis techniques.
This is typically done by analysing text comments or posts on websites and social media platforms to predict a sentiment score \citep{moniz2018multi}.
\citet{moniz2018multi} released a dataset containing 100,000 news items on four topics: \emph{economy}, \emph{microsoft}, \emph{obama} and \emph{palestine} with the respective social feedback on 3 social media platforms: \emph{Facebook}, \emph{Google+} and \emph{LinkedIn}.
Here we attempted a different approach to predict the sentiment score by analysing the number of reactions received for the piece of news on the respective social media platforms. 
We included the \textbf{NewsHeadlineSentiment} and \textbf{NewsTitleSentiment} datasets that aim to predict the sentiment score of news headline and news title using the number of reactions over time from social media platforms illustrated in Figure \ref{fig:news popularity} .

\begin{figure}[!t]
    \centering
    \includegraphics[width=0.75\columnwidth]{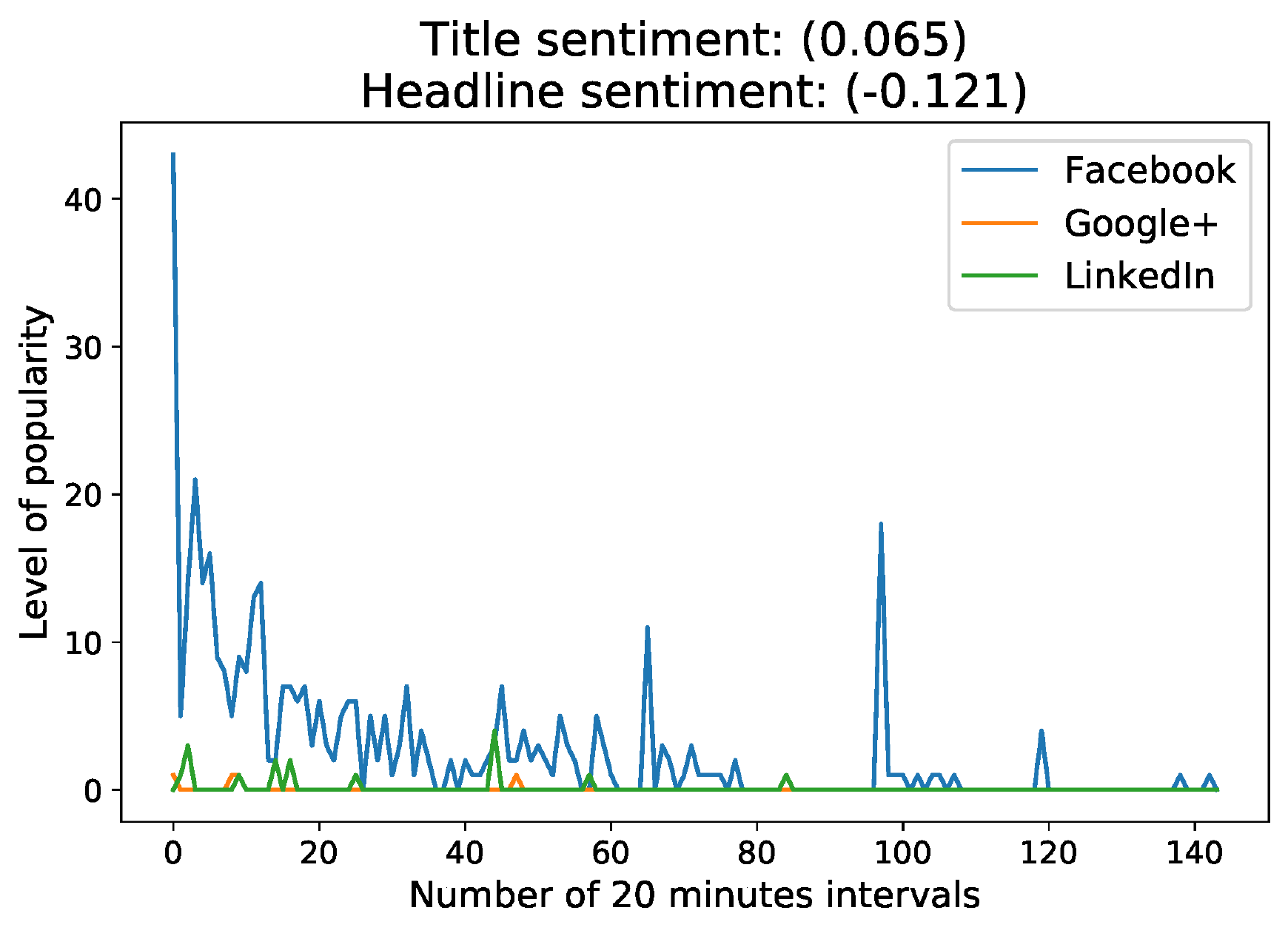}
    \caption{Example of news popularity on 3 social media platforms. The
    title of the news is ``Obama denounces rise of `vulgar and divisive` 	politics of Trump'' with the headline ``And it's worth asking ourselves what each of us may have done to contribute to this vicious atmosphere in our politics,” Obama told the ...'' (\url{https://time.com/4259468/obama-trump-violence-rallies/}). The values in the brackets correspond to the respective sentiment value in news title and headline after 2 days.}
    \label{fig:news popularity}
\end{figure}

\subsubsection{Forecasting}
As described in the introduction and Section \ref{sec:background}, TSF is the task of predicting future values based on some recent and/or seasonal values.
This is usually done by fitting a model to the historical data and extrapolating it into the future.
Our regression problem can be seen as a general case of forecasting where we are still predicting a continuous value that may not necessarily be a future value or depending more heavily on recent values.
Thus, we included in this archive a dataset that could easily be solved with forecasting models to show that forecasting tasks can also be tackled using \ourmethod{} models. 
The \textbf{Covid3Month} dataset contains the daily confirmed number of COVID-19 cases in most countries in the world from January to March 2020, and the goal is to predict the death rate at the start of April 2020. 
An example of the daily confirmed Covid-19 cases and death rate for Italy is shown in Figure \ref{fig:covid}. 

\begin{figure}[!t]
    \centering
    \includegraphics[width=0.85\columnwidth]{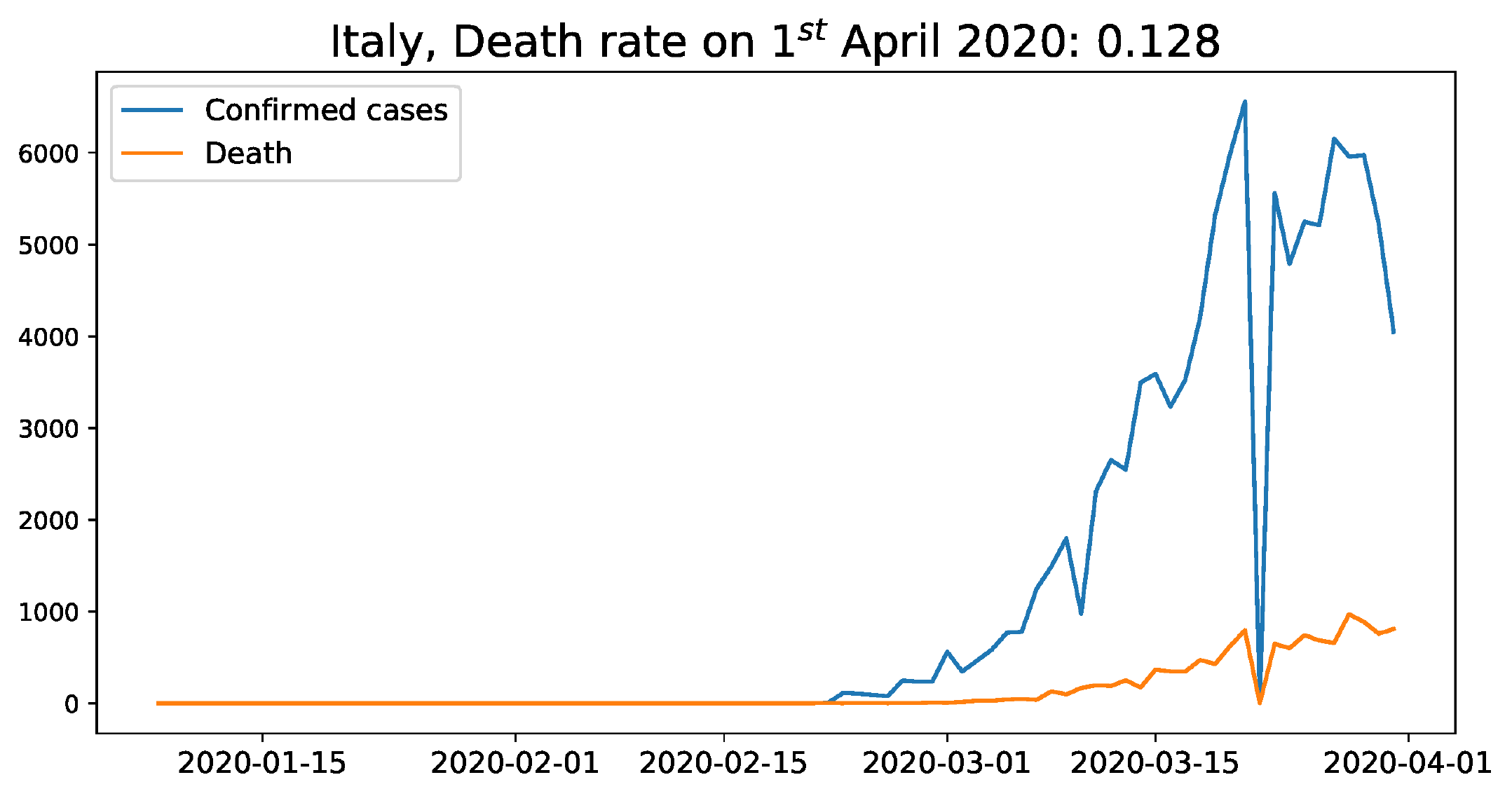}
    \caption{Daily confirmed Covid-19 cases and death rate for Italy.}
    \label{fig:covid}
\end{figure}

\section{Existing algorithms}
\label{sec:algorithms}
In this section, we describe some existing algorithms for \ourmethod{} problems.
Most methods developed in \ourmethod{} cases are highly specific to a problem and are not generalisable, as discussed in Section \ref{subsec:related work}.
We observe the similarity of \ourmethod{} with TSC \citep{bagnall2017great} in Definition \ref{def:tsr}.
The only difference between both tasks is that the target variable is continuous instead of discrete for TSC.
Hence, in principle, most methods developed for TSC can be adapted for \ourmethod{} problems. 
These algorithms are categorised into 4 types: feature-based, dictionary-based, distance-based and deep learning.

\subsection{Feature-based Algorithms}
Feature-based algorithms learn from time series data by extracting discriminating features.
Then these features are used to train a classification or regression model. 
In this section, we discuss some existing feature-based algorithms for time series data.

\subsubsection{Classical regression models}
\label{subsec:classical models}
Classical regression models such as Support Vector Machine (SVM), Linear Regression (LR) and Random Forest (RF) are designed for tabular data.
These models learn a mapping function from some input features extracted from the time series to the target variable. 
A straightforward approach is to treat the time series as tabular data where the time series values are the features. Multidimensional time series will be flattened out into a single long feature vector of length $D \times L$, where $D$ is the number of dimensions in the series and $L$ is the length of the time series.
For instance, a time series with 3 dimensions and 100 data points results in a feature vector with 300 features.
Generally, this approach will not take into account the temporal dimension which is important for discriminating time series because each feature is assumed to be independent of one another.

Despite the simplicity of treating the values of each time series as the features, a more common practice is to extract features from the whole time series. These features are used to characterise the time series which are commonly used for forecasting and visualisation \citep{kang_hyndman_smith-miles_2017,montero-manso_athanasopoulos_hyndman_talagala_2020}. 
The FFORMA algorithm \citep{montero-manso_athanasopoulos_hyndman_talagala_2020} is a feature-based forecast model that trains a meta-model using features extracted from the time series. 
The meta-model is used for assigning weights to various forecasting algorithms based on the characteristics of the time series.
Features are also being used to visualise the performance of forecasting algorithms in an instance space, where time series are represented in a 2-dimensional space \citep{kang_hyndman_smith-miles_2017}. 
These features include the summary statistics of the time series, spectral entropy, trend, seasonality, linearity and autocorrelation are extracted from the time series \citep{kang_hyndman_smith-miles_2017, montero-manso_athanasopoulos_hyndman_talagala_2020}.
The \texttt{tsfeatures} R package\footnote{\url{https://github.com/robjhyndman/tsfeatures}} is a popular package that extracts various features from time series data.
\citet{fulcher_little_jones_2013} introduce the Highly Comparative Time Series Analysis (HCTSA) features set that consists of over 7000 time series features.
The Canonical Time Series Characteristics (Catch22) \citep{lubba_sethi_knaute_schultz_fulcher_jones_2019} is a reduced set of HCTSA that consists of the 22 most discriminating features for TSC, evaluated on the UCR TSC archive.
Although Catch22 when trained with a decision tree classifier is not as accurate as some state-of-the-art TSC algorithms, it is more interpretable, which may important in some applications.

Once the features are extracted, they can be used with any classical regression model. 
Next, we discuss some of the popular regression models.
The SVM \citep{cortes1995support} is a popular classification model.
Support Vector Regression \citep[SVR,][]{drucker1997support} is a variant of SVM designed for regression.
Unlike many regression algorithms that seek to minimize squared error, SVR tries to fit the error rate within a threshold, $\epsilon$ \citep{drucker1997support}. 
SVR works by mapping the data into a higher-dimensional space so that it is linearly separable using a kernel function such as linear, sigmoid or Gaussian Radial Basis Function \citep[RBF,][]{cortes1995support}.
Then it fits a hyperplane through the data bounded by two boundary lines which are $\epsilon$ distance apart from the hyperplane. 
The boundary lines are formed by support vectors which are datapoints that are closest to the boundary.

The RF \citep{breiman2001random} algorithm has proven to be very robust on many classification and regression tasks \citep{segal2004machine}. 
It is a bootstrap aggregation (also known as bagging) ensemble learning method that combines the predictions of multiple decision trees to  improve  prediction accuracy \citep{breiman2001random}.
Bagging is a type of ensemble learning method that randomly samples the data with replacement to build multiple models and aggregates the outputs from all models.
Bagging aims to reduce the variance of high variance models such as decision trees. 
RF builds a multitude of decision trees at training time and outputs the average values of the appropriate leaf for regression tasks \citep{breiman2001random}. 
There are 2 main hyper-parameters that need to be tuned for each problem, the number of trees $N_{tree}$ and the number of features randomly selected at each node $m$ \citep{breiman2001random}. 
One major disadvantage of RF is that it is prone to overfit datasets with noisy classification/regression tasks. 

Extreme Gradient Boosting \citep[XGBoost,][]{chen2016xgboost} is a further accurate and popular machine learning algorithm. 
Similar to RF, XGBoost is a decision tree based ensemble learning algorithm that aims to reduce the variance and bias. 
Different from RF that uses bagging, XGBoost uses gradient boosting with regularisation to avoid overfitting, a problem in RF \citep{chen2016xgboost}.
XGBoost reduces bias by building models sequentially while minimising the errors from previous models \citep{chen2016xgboost}.
The errors are minimised using the gradient descent algorithm.
This essentially ``boosts'' the model's performance over time \citep{chen2016xgboost}. 

\subsubsection{Functional Linear Models}
SoFR is widely studied in the statistics community. Specifically FLM is the most common approach for SoFR as it is simple and intuitive \citep{goldsmith2014estimator}.
FLM extends the standard multiple linear regression model to functional data
\citep{goldsmith2014estimator}. 
It is expressed as $Y_i=\int_0^1 W_i(s)\beta(s)ds + \epsilon_i$, where $Y_i$ is the scalar response, $W_i(s)$ is the functional form of the time series, $\beta(s)$ is the coefficient function and $\epsilon_i$ is the random noise in the data \citep{reiss2017methods,goldsmith2014estimator}}.
Most work in the literature of SoFR had focused on better estimating the $\beta(s)$ coefficient function with various basis functions.

In this work, we will only be focusing on the two most popular basis functions for FLM.
The FPC basis function when applied to FLM is commonly known as functional principal component regression (FPCR).
FPCR is based on functional principal component analysis (FPCA) decomposition \citep{goldsmith2014estimator} that is similar to PCA decomposition where the data is represented by $K_w$ principal components that explain the most variance in the data.
Other than FPC, the smoothness in the coefficient function can be enforced using spline basis functions \citep{goldsmith2014estimator}.
The B-spline basis function is one of the most popular choices where the $\beta(s)$ coefficient function is expressed in terms of $K_B$ B-spline basis.

\subsubsection{Interval-based features}
Instead of extracting features from the whole time series, features can be extracted from the intervals of the time series. It has been shown that these interval-based features generally give better performance than whole series features \citep{deng2013time, bagnall2017great}. 
The Time Series Forest algorithm \citep{deng2013time} is one of the most accurate TSC algorithms \citep{bagnall2017great}.
It extracts three features, mean, standard deviation and slope from an interval of a time series and builds a forest of time series trees, where random intervals are selected in each node of the tree \citep{deng2013time}.

Time series shapelets algorithms \citep{ye2009time,rakthanmanon2013fast,lines2012shapelet}
find the best discriminating shapelets (subsequences) in the data.
The first time series shapelets classifier \citep{ye2009time} trains a decision tree using shapelets as the splitting criterion.
However, the algorithm has very high training complexity as it needs to scan through a high number of shapelet candidates.
Since then, many novel scalable algorithms for shapelet discovery have been proposed 
\citep{rakthanmanon2013fast,mueen2011logical,grabocka2014learning,lines2012shapelet}.
The most accurate shapelet algorithm, Shapelet Transform (ST)\citep{lines2012shapelet}
transforms a time series using the distance of a time series to all $k$ shapelets, creating a feature vector with $k$ dimensions. 
The transformed time series are then used to construct one of the most accurate TSC algorithms, Shapelet Ensemble (SE) \citep{bagnall2015time}.
SE is an ensemble consisting of 8 standard classifiers each applied to the shapelet features.

\subsubsection{Random Convolutional Kernel Transform (Rocket)}
Recently, \citet{dempster2019rocket} proposed the Rocket classifier that achieves state-of-the-art accuracy in TSC with a fraction of the computational expense of existing methods.
Rocket transforms time series using a large number of random convolutional kernels and trains a ridge regression classifier.
These kernels have random length, weights, bias, dilation, and padding, and when applied to a time series produce a feature map.
Then the maximum value and the proportion of positive values are computed from each feature map, producing two real-valued numbers as features per kernel.
With the default 10,000 kernels, Rocket produces 20,000 features.
Rocket was found to be the most accurate TSC classifier compared with other state-of-the-art algorithms such as \hivecote{} \citep{lines2016hive} and InceptionTime \citep{fawaz2019inceptiontime} when benchmarked on the 85 TSC datasets \citep{dau2019ucr}.
As Rocket was designed for classification tasks, in this work, we adapt Rocket for regression tasks by replacing the ridge regression classifier with a ridge regression model. 

\subsection{Dictionary-based Algorithms}
Dictionary-based algorithms transform time series by building a dictionary that represents the observed frequency of a particular pattern or feature in the time series. 
The algorithms then learn to discriminate between different time series by comparing the dictionary of the two time series.
This is also known as the ``bag of words'' algorithm where the patterns (subsequences) are discretized into words.

There are various bag of words algorithms for TSC.
Notably some of the popular ones are the Bag of Patterns (BOP) \citep{lin2012rotation}, Symbolic Aggregation Approximation Vector Space (SAXVSM) \citep{senin2013sax}, Bag of Symbolic Fourier Approximation (SFA) Words (BOSS) \citep{schafer2015boss}, Word  Extraction  for TSC (WEASEL) \citep{schafer2017fast} and WEASEL + Multivariate Unsupervised Symbols and Derivatives (MUSE) \citep{schafer2017multivariate}.

The recent TSC benchmark survey \citep{bagnall2017great} ranks BOSS as the most accurate dictionary-based classifier. 
BOSS builds a dictionary using SFA words \citep{schafer2015boss}. 
Each subsequence in the time series is transformed into SFA words using truncated discrete fourier transform, making it robust to noise. 

Although the survey \citep{bagnall2017great} did not compare with WEASEL, WEASEL is arguably more accurate than BOSS \citep{schafer2017fast}.
WEASEL improves on BOSS by determining discriminative Fourier coefficients using ANOVA f-test and applying Chi-Squared test for feature selection \citep{schafer2017fast}.
WEASEL+MUSE aims to tackle multivariate TSC by splitting the time series into its dimensions and applying the univariate transformation to each dimension \citep{schafer2017multivariate}.
It also transforms the derivative of each dimension into words and concatenates these with a dimension identifier to enrich the feature space. 
Finally, similar to WEASEL, a feature selection technique is applied to filter out non-discriminative features \citep{schafer2017multivariate}.

\subsection{Distance-based Algorithms}
The majority of TSC research has been focused on the similarity of two time series.
This involves matching two time series and computing the distance between them. 
Then, a $k$-nearest neighbour ($k$-NN) algorithm is applied to find the most similar time series \citep{lines2015time,tan2020fastee}.

The $k$-Nearest Neighbour ($k$-NN) algorithm is one of the simplest and most intuitive algorithms \citep{sammut2011encyclopedia}.
A $k$-NN algorithm requires two parameters, (1) the number of nearest neighbours $k$ and (2) a distance metric \citep{sammut2011encyclopedia}. 
Similar to any other classical regression models described in Section \ref{subsec:classical models}, $k$-NN was initially designed for tabular data.
Some examples of distance metrics for tabular data are the Euclidean, Manhattan, Minkowski and Mahalanobis distances.
Using one of these distance metrics, 
the model finds $k$ nearest instances from the training dataset to a query instance in the feature space \citep{sammut2011encyclopedia}.
For regression, the target values of the $k$ nearest neighbours are averaged out and assigned as the target of the query instance. 
Weighted average can also be applied using the distances to the query to put more emphasis on nearer neighbours.  

For time series data, the $k$-NN algorithm has to take into account the temporal dimension of the data.
Hence, the distance measures \citep{lines2015time,tan2020fastee} are also different from classic $k$-NN algorithms for tabular data. 
They are commonly known as elastic distances.
\textcolor{black}{The simplest is the Euclidean distance (ED), 
which is similar to the ED used for tabular data.
Equation \ref{eqn:ed} describes the ED to compute the distance between two time series $P$ and $Q$, where $D$ is the number of dimensions and $L$ is the length of the time series. 
\begin{equation}
	ED(P, Q) = \sum_{j=1}^{D}\sqrt{\sum_{i=1}^{L}(p_i^j-q_i^j)^2}
	\label{eqn:ed}
\end{equation}
A limitation of ED is that it cannot allow for processes that are not directly aligned or which unfold at differing rates.}  

\begin{figure}
	\centering
	\begin{subfigure}{0.49\linewidth}
		\includegraphics[width=\linewidth]{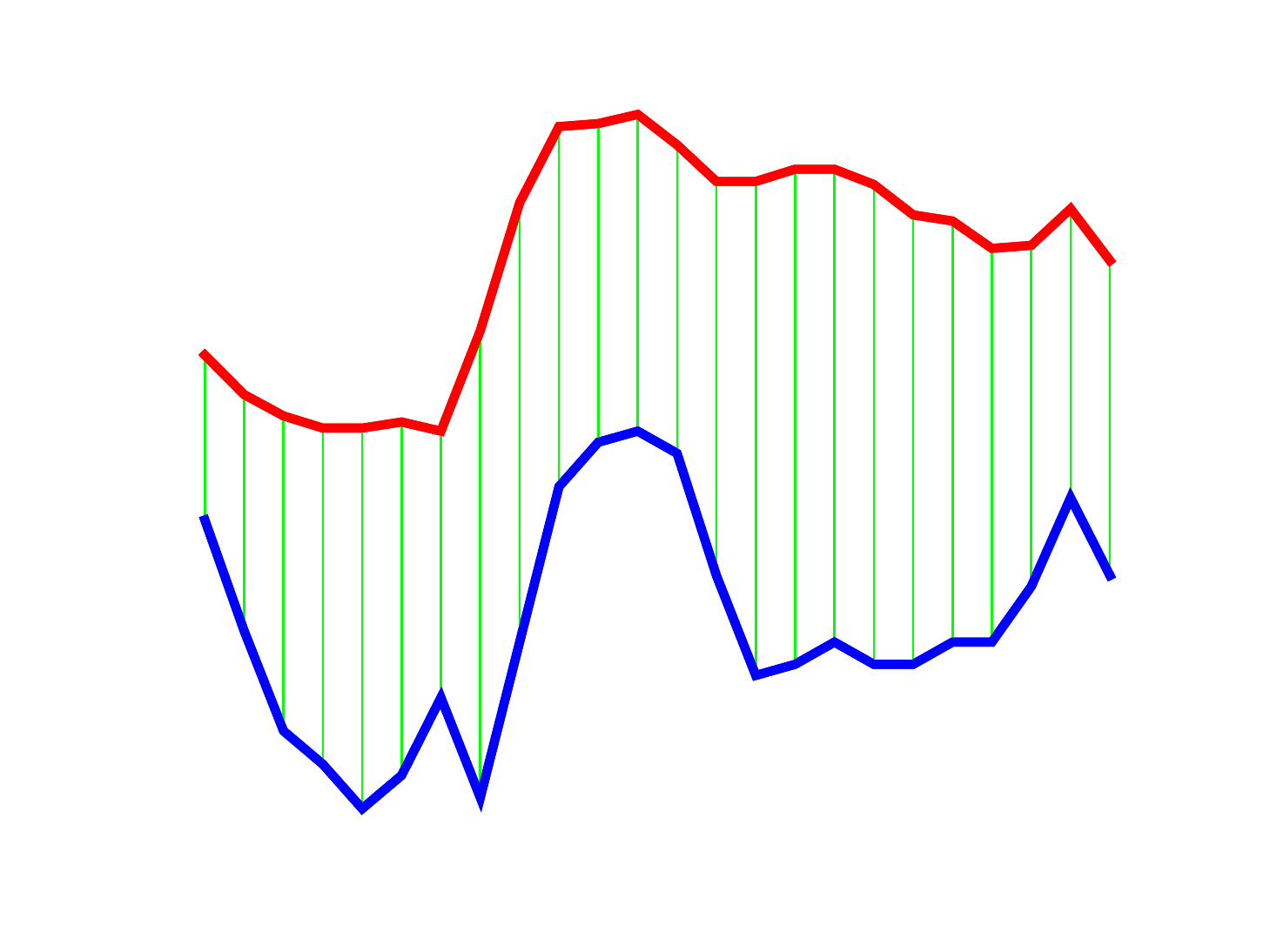}
		\caption{}
		\label{fig:ed}
	\end{subfigure}
	\begin{subfigure}{0.49\linewidth}
		\includegraphics[width=\linewidth]{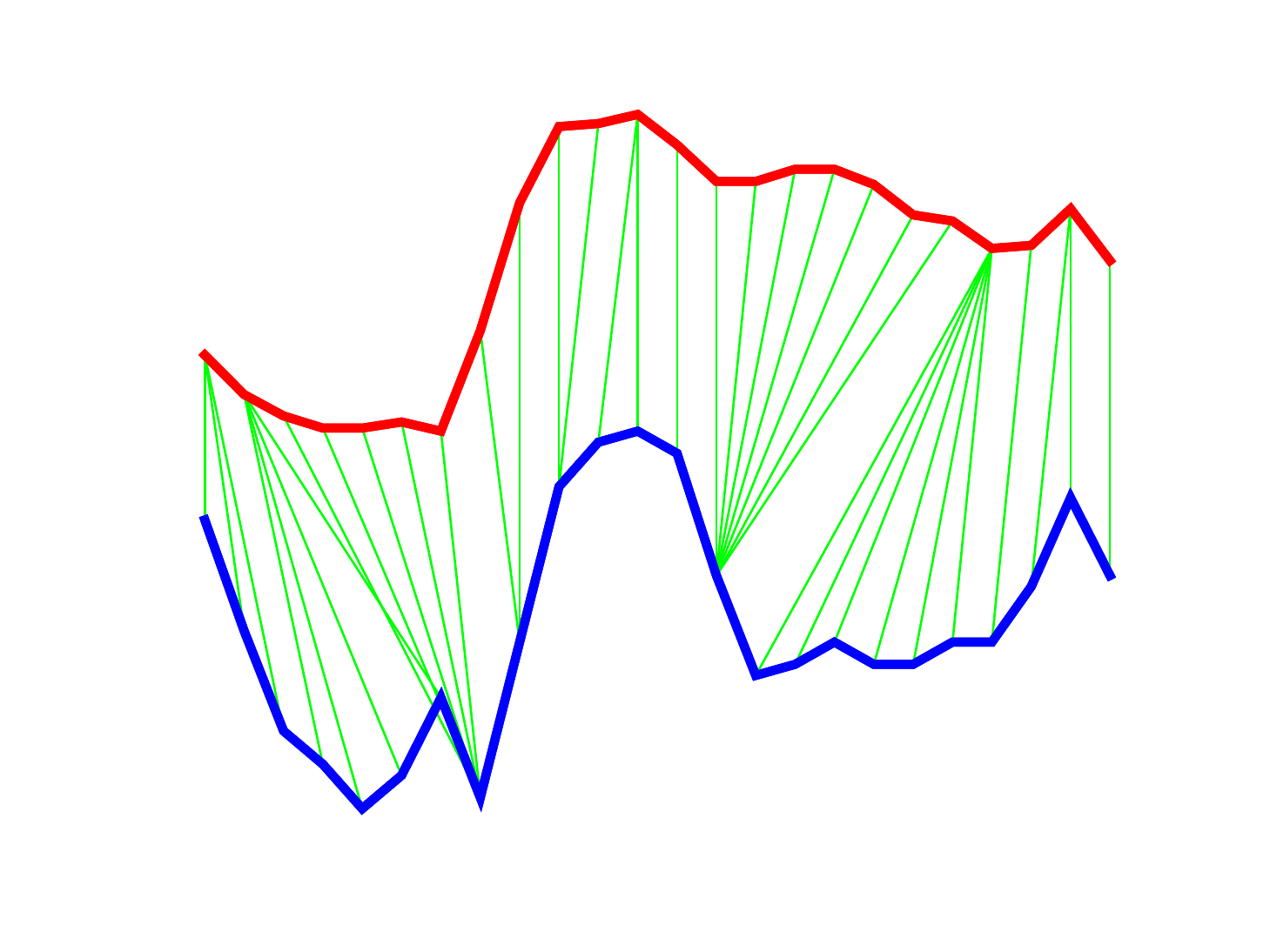}
		\caption{}
		\label{fig:dtw}
	\end{subfigure}
	\caption{Example of alignment of two time series using (a) Euclidean distance and (b) DTW distance}
\end{figure}

\textcolor{black}{Distance measures that do make such allowance are known as \textit{elastic distances}. One popular example is the Dynamic Time Warping (DTW) distance.}
DTW computes the minimum distance of two time series by finding the optimum alignment of two time series and taking into account the temporal order of the data \citep{lines2015time,tan2020fastee, tan2018efficient}.
\textcolor{black}{Typically, DTW is computed with a warping constraint that limits the warping path \citep{tan2018efficient}. 
	This has the effect of minimising irregular warping and reducing the time complexity of DTW \citep{tan2018efficient, tan2020fastee}.
	Since DTW is a widely studied distance measure, we refer interested readers to the following papers \citep{tan2018efficient, tan2020fastee} for more details. 
}
Figure \ref{fig:ed} and \ref{fig:dtw} illustrate the differences between ED and DTW distance.
For multivariate time series, DTW can be computed dependent or independent of the dimensions of the time series, commonly known as $DTW_D$ and $DTW_I$ \citep{shokoohi2017generalizing}. 

There are various other distance measures other than ED and DTW, \textcolor{black}{none of which dominates the others in terms of classification accuracy, but each of which excels at some tasks} \citep{lines2015time}. 
The Ensembles of Elastic Distances (EE) \citep{lines2015time} is a combination of 11 elastic distances that is significantly more accurate than each of the individual distances. 
Although accurate, EE is not computationally efficient as it requires a grid search over a range of parameters for each elastic distance. 
FastEE \citep{tan2020fastee} is a significantly faster version of EE that trims the parameter space by leveraging off the properties of each elastic distances. Instead of performing a grid search, Proximity Forest (PF) \citep{lucas2019proximity} is a tree-based algorithm where an elastic distance and its parameters are selected at random at each node of the tree. PF has shown to be significantly more accurate and faster than EE for many TSC tasks \citep{lucas2019proximity}. 
Although the modification of the NN algorithm for regression tasks is very straightforward, applying EE or PF to regression tasks requires more complex modification of the algorithm which we leave for future work.
In this work, we focus only on the two most popular TSC NN algorithms, NN with ED (NN-ED) and DTW distance (NN-DTW).

\subsection{Deep Learning Algorithms}
Deep learning models are capable of predicting both discrete labels (classification) and continuous values (regression).
Fundamentally, the output of a neural network is a continuous value.  
Typically for classification tasks, softmax activation is used at the output layer to output class probabilities and classification is done by taking the class with the highest probability. 
The softmax activation is replaced with linear activation for regression tasks. 
Apart from the activation functions, the loss function has to be changed as well. 
The categorical cross entropy loss function that is commonly used for classification can be replaced by either the mean squared error or the mean absolute error loss function for regression tasks, in this case, mean squared error is chosen. 
Recently, several deep learning models have been developed and benchmarked for TSC \citep{fawaz2019deep, wang2017time,fawaz2018transfer,fawaz2019inceptiontime}.
In this work, we adapted three TSC deep learning models, Residual Networks (ResNet), Fully Convolutional Neural Networks (FCN) and Inception network \citep{fawaz2019inceptiontime}. 

ResNet and FCN were first proposed in \citep{wang2017time}.
In a recent survey on deep learning for TSC \citep{fawaz2019deep},
ResNet was ranked the most accurate univariate TSC model benchmarked on 85 univariate time series datasets \citep{dau2019ucr}. 
ResNet consists of 3 residual blocks with 3 convolutional layers in each block, followed by a global average pooling layer and an output layer.
Different from the typical convolutional networks, ResNet has a shortcut residual connection between the convolutional layers which makes training easier by reducing the vanishing gradient effect.

FCN is the most accurate deep learning model for multivariate TSC on 12 multivariate time series datasets \citep{baydogan2015learning} and the second most accurate deep learning model for univariate TSC.
It is composed of three convolutional blocks with batch normalisation and a ReLU activation function. 
Then, global average pooling is applied to the last convolutional block and connected to a softmax classifier \citep{fawaz2019deep}.
For regression, the softmax activation function is replaced with linear activation function. 

\citet{fawaz2019inceptiontime} recently proposed the Inception network, which significantly improved existing deep learning models and achieved competitive performance with the state-of-the-art TSC model, HIVE-COTE \citep{lines2016hive}.
The Inception network consists of two different residual blocks connecting the input to the next block's input to mitigate the vanishing gradient problem \citep{fawaz2019inceptiontime}.
Each residual block is comprised of three Inception modules.
There are two major components in each of the inception module.
The first one is the bottleneck layer that reduces the dimension of the time series using $m$ filters and also allowing the Inception network to have ten times longer filters than ResNet \citep{fawaz2019inceptiontime}.
The second component consists of sliding multiple filters of different lengths to the output of the first component.
A MaxPooling operation is also applied to the time series in parallel to these two components. 
The output from each of the convolution and MaxPooling operation is then concatenated to form the output of the Inception module.
Finally, global average pooling is applied to the final residual block and passed to a fully connected layer for classification. 

In our work, we use the same architecture from the original papers \citep{fawaz2019deep,fawaz2019inceptiontime} with some minor modifications to the activation and loss functions as mentioned above. 
We refer interested readers to the respective papers for the details of these architectures.

\section{Benchmarking results}
\label{sec:results}
In this section, we evaluate the regression algorithms described in Section \ref{sec:algorithms} and set a baseline using the datasets from our \ourmethod{} archive \citep{tan2020monash} described in Section \ref{sec:datasets}.
We evaluate and benchmark the following regression algorithms:
\begin{enumerate}
    \item FPCR \citep{goldsmith2014estimator}
    \item FPCR with B-spline \citep{goldsmith2014estimator}
    \item Grid-search optimised SVR \citep{drucker1997support}
    \item RF \citep{breiman2001random} with 100 trees
    \item XGBoost \citep{chen2016xgboost} with 100 trees
    \item NN-ED with $k=1,5$ (1-NN-ED and 5-NN-ED)
    \item NN-DTW with $k=1,5$ (1-NN-DTW and 5-NN-DTW) and warping window $w=0.1\times L$
    \item FCN \citep{fawaz2019deep}
    \item ResNet \citep{fawaz2019deep}
    \item Inception Network \citep{fawaz2019inceptiontime}
    \item Rocket \citep{dempster2019rocket}
\end{enumerate}
Missing values in the time series are linearly interpolated. 
When using a traditional regression model (i.e. non-temporal), the time series are flattened out into a single long feature vector.

We used the standard Scikit-Learn Python library \citep{scikit-learn} to implement SVR and RF algorithms. 
The SVR algorithm is optimised by performing a 3-fold cross validation with grid search on the hyper-parameters.
Specifically, the kernel, gamma and $C$ parameters are optimised from a standard range of values. 
The kernel function is selected from RBF and Sigmoid. 
The gamma parameter selected from $[0.001, 0.01, 0.1, 1]$ defines the influence of support vectors. 
The regularisation parameter $C$ is selected from $[0.1, 1, 10, 100]$. XGBoost was implemented using the Python XGBoost library\footnote{\url{https://xgboost.readthedocs.io/en/latest/python/python_intro.html}}.
Apart from the number of trees, we use the default parameters for both RF and XGBoost from the Python libraries. 
Our empirical experiments show that RF and XGBoost with parameters optimised using a grid search strategy performs similarly or worse than the default parameters and takes a significantly longer time to train. 
Hence they are excluded from this benchmark.
The FPCR and FPCR with B-spline models are implemented using the Scikit-FDA Python package\footnote{\url{https://fda.readthedocs.io/en/latest/}}, a library for functional data analysis in Python.

For time series algorithms, we adapted the code from \citet{fawaz2019deep}\footnote{\url{https://github.com/hfawaz/dl-4-tsc}} for both ResNet and FCN and \citet{fawaz2019inceptiontime}\footnote{\url{https://github.com/hfawaz/InceptionTime}} for Inception Network.
The code for Rocket was taken from \citet{dempster2019rocket}\footnote{\url{https://github.com/angus924/rocket}} and modified for multivariate time series with the help from the original authors. 
The multivariate version of Rocket applies the transformation to each dimension independently.

The time series NN algorithms were all implemented in Java.
Our source code has been made open source online at \url{https://github.com/ChangWeiTan/TS-Extrinsic-Regression}. 

Since some of the algorithms are non-deterministic, we evaluate all the algorithms over 5 runs and report the average root mean squared error (RMSE), one of the most widely used metrics for regression tasks. 
Equation \ref{eqn:rmse} describes the formal definition of RMSE where $n$ is the number of instances, $y_i$ and $\hat{y_i}$ are the actual and predicted target respectively.

\begin{equation}
    RMSE = \sqrt{\frac{\sum_{i=1}^n (\hat{y_i}-y_i)^2}{n}}
    \label{eqn:rmse}
\end{equation}

We compare the algorithms statistically over the current datasets following the recommendations from \citep{demvsar2006statistical}.
First, we rank each algorithm by RMSE for every dataset.
Rank 1 is assigned to the algorithm with the lowest RMSE while rank 13 is assigned to the highest one.
Fractional ranking is assigned to the algorithm in case of ties.
We then compute the average rank for each algorithm.
Then, the Friedman test \citep{friedman1940comparison,demvsar2006statistical} was applied to the average ranks.
If the null hypothesis is rejected, the post-hoc two-tailed Nemenyi test is used to compare the algorithms to each other \citep{demvsar2006statistical}.
Using this test, the performance of the algorithms is significantly different if the average ranks differ by at least the critical difference shown in Equation \ref{eqn:cd}, where $q_{\alpha}=3.313$ is the critical value for $\alpha=0.05$, $k=13$ being the number of algorithms and $N=19$ being the number of datasets.
This gives $CD=4.186$.

\begin{equation}
    CD = q_{\alpha}\sqrt{\frac{k(k+1)}{6N}}
    \label{eqn:cd}
\end{equation}

\begin{figure}
    \centering
    \includegraphics[width=0.85\columnwidth]{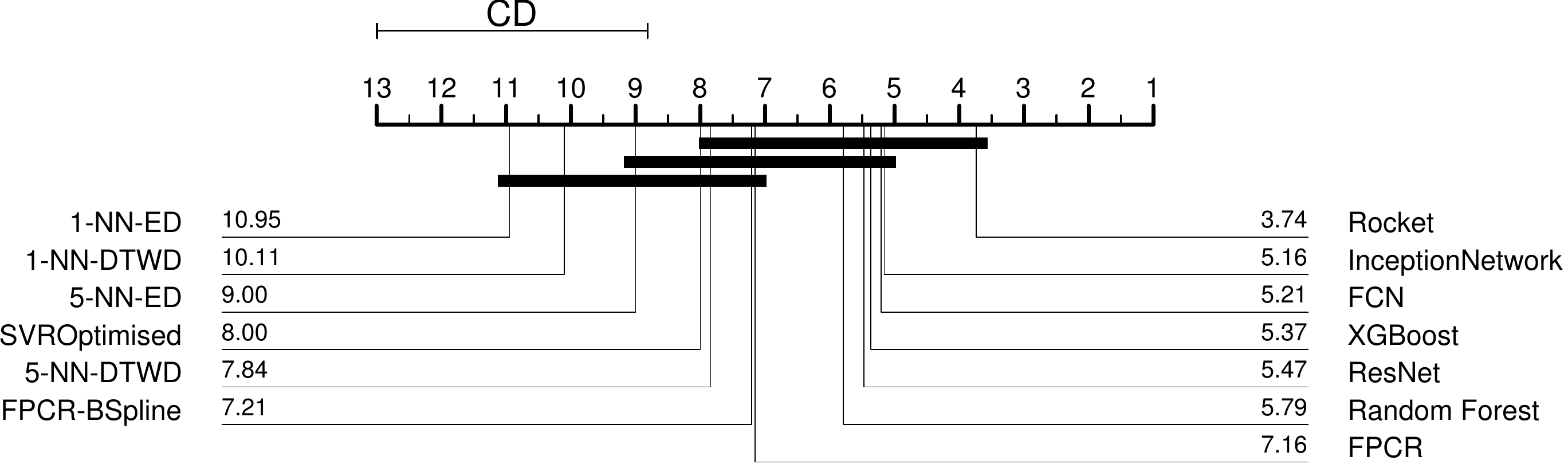}
    \caption{Critical difference diagram showing statistical difference comparison of 13 regression algorithms on the current regression archive}
    \label{fig:cd}
\end{figure}

Finally, a critical difference diagram was used to visualise the comparison, where the thick horizontal line connecting a group of algorithms indicates that all the algorithms in the group are not significantly different from one another \citep{demvsar2006statistical}. 
Figure \ref{fig:cd} shows the critical difference diagram of comparing the algorithms used to benchmark the existing archive.
The average ranks are indicated next to the algorithms in the figure. 

The critical difference diagram in Figure \ref{fig:cd} shows that Rocket is the most accurate algorithm with an average rank of $3.74$ and is significantly different from 1-NN-ED and 1-NN-DTWD.
The figure also shows that there is no significant difference between the state-of-the-art time series algorithms and classical regression algorithms.
However, our experiments indicate that Rocket is the most computationally efficient compared to all other algorithms. 

We further compare the relative performance of each algorithm on the current \ourmethod{} archive.
The relative performance of an algorithm is computed by scaling the RMSE of each algorithm with the median RMSE obtained for a dataset. 
Equation \ref{eqn:scaled rmse} describes the equation to scale the RMSE of algorithm $j$ for dataset $i$. 

\begin{equation}
    scaled\_RMSE_i^j = \frac{RMSE_i^j}{RMSE_i^j + median(RMSE_i)}
    \label{eqn:scaled rmse}
\end{equation}

\begin{figure}
    \centering
    \includegraphics[width=0.8\columnwidth]{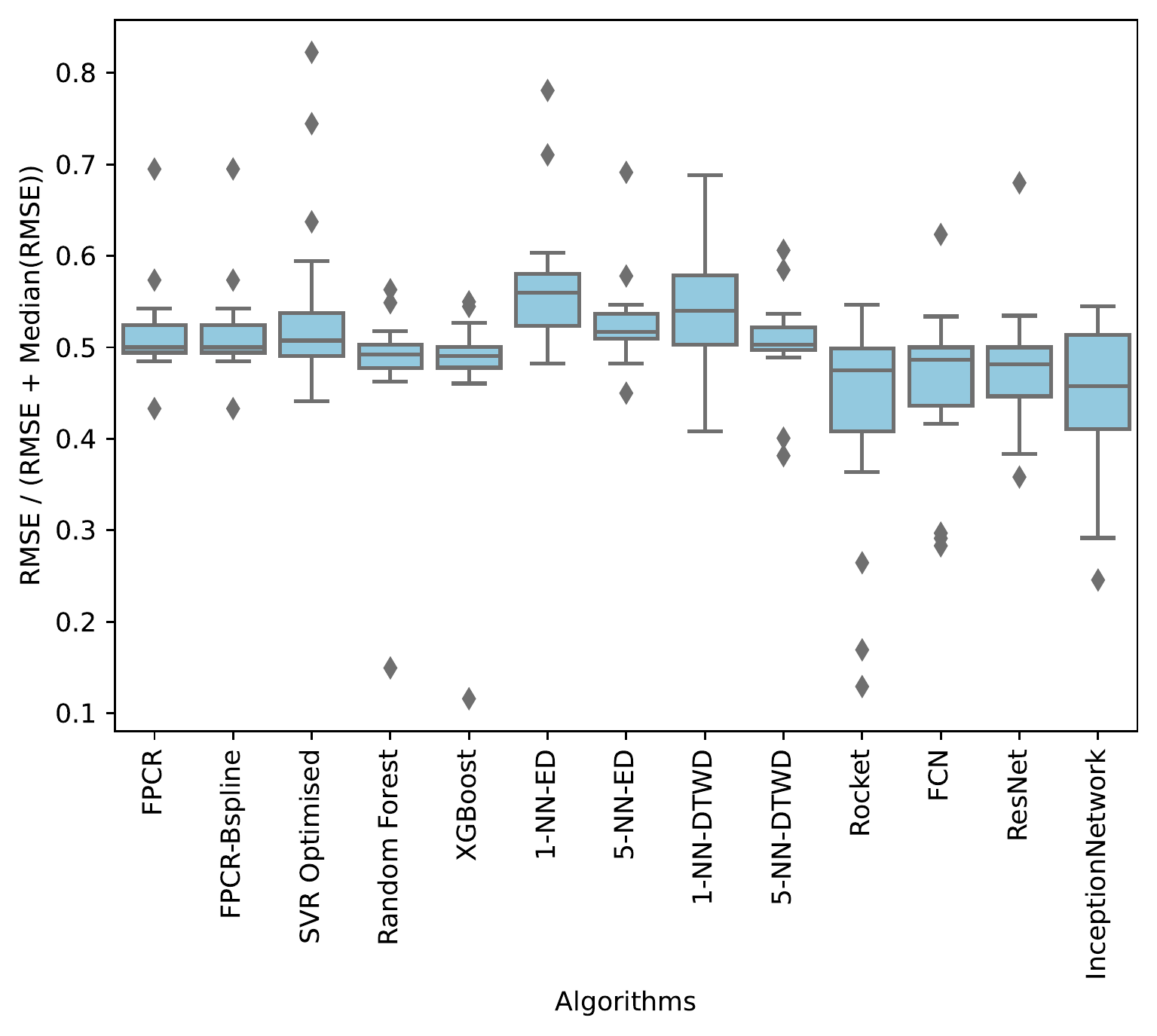}
    \caption{The relative RMSE of each algorithm on the current \ourmethod{} archive. Values greater than 0.5 indicate that the algorithm has RMSE higher than the average algorithm while values smaller than 0.5 indicate an RMSE lower than the average algorithm.}
    \label{fig:boxplot}
\end{figure}

The algorithm with median RMSE can be interpreted as the algorithm that gives the average performance for the dataset. 
Hence, values larger than 0.5 indicate worse performance, while values smaller than 0.5 indicate a better performance than the average performance.
Figure \ref{fig:boxplot} illustrates the scaled RMSE for each algorithm in the form of boxplots. 
It shows that most algorithms have their values around 0.5,
while bespoke time series algorithms such as Rocket, FCN, ResNet and Inception Network have larger spread in the values and tend to have values smaller than 0.5.
This implies that when time series algorithms perform better, they perform significantly better than the other algorithms, while the other algorithms tend to perform similarly to an average algorithm. 
The median of all algorithms are similar, around 0.5, which suggests that there is room for better algorithms to be developed for \ourmethod{} problems.

Table~\ref{tab:results} shows a more detailed results of these algorithms on all the datasets in the archive.
The results show that Rocket performs the best overall with the lowest average RMSE ranks followed by the other state-of-the-art TSC algorithms.
RF and XGBoost are both very competitive compared with the time series algorithms. 
This is expected as XGBoost and RF are both the state of the art in ML algorithms, especially in popular data science and ML competitions \citep{nielsen2016tree}. 
The results also indicate that the SoFR algorithms are also competitive as they are not significantly different from the standard regression algorithms.
This further strengthens our findings from Figure \ref{fig:boxplot} that there is room for better algorithms to be developed for \ourmethod{} problems and that new algorithms should also be computationally efficient.

On the tasks of energy and health monitoring, time series algorithms are clearly performing better than classical regression algorithms, with the top 3 algorithms being time series algorithms.
For instance, Inception network performs the best on heart rate prediction tasks while Rocket is the most accurate on energy prediction tasks.
There is no clear winner for environment monitoring tasks.
Classical regression algorithms perform better at predicting pollution level while time series algorithms perform better on the remaining datasets. 
The reason is that, the pollution metrics from these pollution datasets can be estimated fairly easily by applying a threshold to the measurements from gas sensors, where classical regression algorithms such as RF and XGBoost are very good at.
Nonetheless, we expect a \ourmethod{} algorithm that uses feature extraction techniques such as the TSC counterparts, Shapelet Transform \citep{lines2012shapelet}, Time Series Forest \citep{deng2013time} and BOSS \citep{schafer2015boss}, will perform better than classical regression algorithms. 

Although there is also no clear winner on the new sentiment analysis task that we propose in this work, the results show that predicting sentiment scores using time series data is feasible, achieving very low RMSE scores.
Both classical regression and time series algorithms perform similarly on forecasting tasks.
This is expected as both types of algorithms are not designed for forecasting and we expect that a forecasting algorithm if adapted for \ourmethod{} will perform better. 
Besides, the small Covid3Month dataset with 140 time series of length 84 may not have enough data for the algorithms to train on. 
Overall, the results indicate that there is a need to design better \ourmethod{} algorithms that can better generalise for most datasets.

\begin{sidewaystable}[]
	\centering\vspace*{12cm}
	\resizebox{\textwidth}{!}{
		\begin{tabular}{l|c|c|c|c|c|c|c|c|c|c|c|c|c}
			\hline
			& \multicolumn{11}{c}{RMSE} \\ 
			\hline
			Dataset & FPCR & FPCR-BSpline & SVR Optimised & Random Forest & XGBoost & 1-NN-ED & 5-NN-ED & 1-NN-DTWD & 5-NN-DTWD & Rocket & FCN & ResNet & Inception \\
			\hline
			AppliancesEnergy & 5.41 & 5.41 & 3.46 & 3.46 & 3.49 & 5.23 & 4.23 & 6.04 & 4.02 & \textbf{2.30} & 2.87 & 3.07 & 4.44 \\
			HouseholdPowerConsumption1 & 147.55 & 147.55 & 152.39 & 248.86 & 231.09 & 473.93 & 432.60 & 427.04 & 297.22 & \textbf{132.80} & 162.24 & 193.21 & 153.72 \\
			HouseholdPowerConsumption2 & 46.93 & 46.93 & 55.98 & 46.93 & 44.37 & 71.48 & 64.27 & 58.80 & 51.50 & \textbf{32.61} & 46.83 & 39.08 & 39.41 \\
			\hline
			BenzeneConcentration & 11.09 & 11.10 & 4.79 & 0.86 & \textbf{0.64} & 6.54 & 5.85 & 4.98 & 4.87 & 3.36 & 4.99 & 4.06 & 1.59 \\
			BeijingPM10Quality & 99.73 & 99.73 & 110.57 & 94.07 & \textbf{93.14} & 139.23 & 115.67 & 139.14 & 115.50 & 120.06 & 94.35 & 95.49 & 96.75 \\
			BeijingPM25Quality & 69.38 & 69.37 & 75.73 & 63.30 & \textbf{59.50} & 88.19 & 74.16 & 88.26 & 72.72 & 62.77 & 59.73 & 64.46 & 62.23 \\
			LiveFuelMoistureContent & 37.68 & 37.68 & 39.73 & 32.16 & 32.44 & 47.84 & 38.54 & 39.97 & 35.19 & 29.41 & 33.26 & 30.35 & \textbf{28.80} \\
			FloodModeling1 & 0.02 & 0.02 & 0.05 & 0.02 & 0.02 & 0.02 & 0.02 & 0.01 & 0.01 & \textbf{0.00} & 0.01 & 0.01 & 0.02 \\
			FloodModeling2 & 0.02 & 0.02 & 0.08 & \textbf{0.01} & 0.02 & 0.02 & 0.02 & 0.02 & 0.02 & \textbf{0.01} & \textbf{0.01} & \textbf{0.01} & \textbf{0.01} \\
			FloodModeling3 & 0.02 & 0.02 & 0.04 & 0.02 & 0.02 & 0.02 & 0.02 & 0.01 & 0.01 & \textbf{0.00} & 0.01 & 0.02 & 0.01 \\
			AustraliaRainfall & 8.44 & 8.44 & 8.65 & 8.39 & 8.49 & 30.25 & 10.23 & 12.00 & 11.95 & \textbf{8.12} & 8.43 & 8.18 & 8.84 \\
			\hline
			PPGDalia & 20.67 & 20.67 & 19.01 & 17.53 & 16.58 & 21.88 & 18.28 & 26.03 & 20.77 & 14.05 & 13.04 & 11.38 & \textbf{9.92} \\
			IEEEPPG & 31.38 & 31.38 & 37.25 & 32.11 & 31.49 & 33.21 & 27.11 & 37.14 & 33.57 & 36.52 & 34.33 & 33.15 & \textbf{23.90} \\
			BIDMC32HR & 13.98 & 13.98 & 13.39 & 15.02 & 13.96 & 14.84 & 14.76 & 15.29 & 15.13 & 13.94 & 13.13 & 10.74 & \textbf{9.43} \\
			BIDMC32RR & 3.37 & 3.37 & 3.17 & 4.35 & 4.37 & 4.39 & 4.14 & 3.53 & 3.43 & 4.09 & 3.58 & 3.92 & \textbf{3.02} \\
			BIDMC32SpO2 & 4.95 & 4.95 & 4.80 & 4.57 & \textbf{4.45} & 5.53 & 5.41 & 5.22 & 5.12 & 5.22 & 5.97 & 5.99 & 5.58 \\
			\hline
			NewsHeadlineSentiment & \textbf{0.14} & \textbf{0.14} & \textbf{0.14} & 0.15 & \textbf{0.14} & 0.20 & 0.16 & 0.20 & 0.16 & \textbf{0.14} & 0.15 & 0.15 & 0.15 \\
			NewsTitleSentiment & \textbf{0.14} & \textbf{0.14} & \textbf{0.14} & \textbf{0.14} & \textbf{0.14} & 0.19 & 0.15 & 0.19 & 0.15 & \textbf{0.14} & \textbf{0.14} & \textbf{0.14} & 0.16 \\
			\hline
			Covid3Month & 0.05 & 0.05 & 0.07 & \textbf{0.04} & 0.05 & 0.05 & \textbf{0.04} & 0.05 & \textbf{0.04} & \textbf{0.04} & 0.07 & 0.10 & 0.05 \\
			\hline
			Average rank & 7.16 & 7.21 & 8.00 & 5.79 & 5.37 & 10.95 & 9.00 & 10.11 & 7.84 & \textbf{3.74} & 5.21 & 5.47 & 5.16 \\
			\hline
		\end{tabular}
	}
	\caption{RMSE obtained for the different algorithms on the \ourmethod{} archive. The lowest RMSE per dataset is indicated in bold.}
	\label{tab:results}
\end{sidewaystable}

\vspace{-10pt}
\section{Conclusion and Future Work}
\label{sec:conclusion}
In this paper, we introduced and motivated the \emph{Time Series Extrinsic Regression} problem where the goal is to predict a continuous value using time series data.
We showed some examples of real-life applications where \ourmethod{} may be useful and discussed some existing methods for this task.
We benchmarked these methods on the first \ourmethod{} benchmarking archive and showed that Rocket, one of the state-of-the-art TSC algorithms performs the best overall. 
Although time series specific Rocket achieved the highest overall rank on accuracy, its rank is not statistically distinguishable from classical machine learning algorithms XGBoost and Random Forest that ignore the temporal order of the data. This is in contrast to the state-of-the-art in time series classification, where bespoke algorithms significantly outperform approaches that ignore the temporal information in the data.
Therefore, this suggests much research is needed to develop better algorithms to improve accuracy on \ourmethod{} problems.

\vspace{-10pt}
\section*{Acknowledgement}
\vspace{-2pt}
This research has been supported by Australian Research Council grant
DP210100072; and the Air Force Office of Scientific Research, Asian Office of Aerospace Research and Development (AOARD) under award number FA2386–18–1–4030.
The authors appreciate the data donation from all the donors and would like to thank the authors of \citet{fawaz2019deep} and \citet{dempster2019rocket} for providing their source code online.

\bibliographystyle{spbasic}
\bibliography{references}

\begin{thebibliography}{66}
\providecommand{\natexlab}[1]{#1}
\providecommand{\url}[1]{{#1}}
\providecommand{\urlprefix}{URL }
\expandafter\ifx\csname urlstyle\endcsname\relax
  \providecommand{\doi}[1]{DOI~\discretionary{}{}{}#1}\else
  \providecommand{\doi}{DOI~\discretionary{}{}{}\begingroup
  \urlstyle{rm}\Url}\fi
\providecommand{\eprint}[2][]{\url{#2}}

\bibitem[{Bagnall et~al.(2015)Bagnall, Lines, Hills, and
  Bostrom}]{bagnall2015time}
Bagnall A, Lines J, Hills J, Bostrom A (2015) Time-series classification with
  {COTE}: The collective of transformation-based ensembles. {IEEE} Transactions
  on Knowledge and Data Engineering 27(9):2522--2535

\bibitem[{Bagnall et~al.(2017)Bagnall, Lines, Bostrom, Large, and
  Keogh}]{bagnall2017great}
Bagnall A, Lines J, Bostrom A, Large J, Keogh E (2017) The great time series
  classification bake off: a review and experimental evaluation of recent
  algorithmic advances. Data Mining and Knowledge Discovery 31(3):606--660

\bibitem[{Baydogan and Runger(2015)}]{baydogan2015learning}
Baydogan MG, Runger G (2015) Learning a symbolic representation for
  multivariate time series classification. Data Mining and Knowledge Discovery
  29(2):400--422

\bibitem[{Box and Jenkins(1970)}]{box1970time}
Box GE, Jenkins GM (1970) Time series analysis forecasting and control. Tech.
  rep., Wisconsin University, Dept of Statistics

\bibitem[{Breiman(2001)}]{breiman2001random}
Breiman L (2001) Random forests. Machine learning 45(1):5--32

\bibitem[{Chatfield(1978)}]{chatfield1978holt}
Chatfield C (1978) The {Holt-Winters} forecasting procedure. Journal of the
  Royal Statistical Society: Series C (Applied Statistics) 27(3):264--279

\bibitem[{Chen and Guestrin(2016)}]{chen2016xgboost}
Chen T, Guestrin C (2016) {XGBoost}: A scalable tree boosting system. In:
  Proceedings of the 22nd {ACM} {SIGKDD} International Conference on Knowledge
  Discovery and Data Mining, pp 785--794

\bibitem[{Cortes and Vapnik(1995)}]{cortes1995support}
Cortes C, Vapnik V (1995) Support-vector networks. Machine learning
  20(3):273--297

\bibitem[{Dau et~al.(2019)Dau, Bagnall, Kamgar, Yeh, Zhu, Gharghabi,
  Ratanamahatana, and Keogh}]{dau2019ucr}
Dau HA, Bagnall A, Kamgar K, Yeh CCM, Zhu Y, Gharghabi S, Ratanamahatana CA,
  Keogh E (2019) The {UCR} time series archive. IEEE/CAA Journal of Automatica
  Sinica 6(6):1293--1305

\bibitem[{De~Vito et~al.(2008)De~Vito, Massera, Piga, Martinotto, and
  Di~Francia}]{de2008field}
De~Vito S, Massera E, Piga M, Martinotto L, Di~Francia G (2008) On field
  calibration of an electronic nose for benzene estimation in an urban
  pollution monitoring scenario. Sensors and Actuators B: Chemical
  129(2):750--757

\bibitem[{Dempster et~al.(2020)Dempster, Petitjean, and
  Webb}]{dempster2019rocket}
Dempster A, Petitjean F, Webb GI (2020) {ROCKET}: exceptionally fast and
  accurate time series classification using random convolutional kernels. Data
  Mining and Knowledge Discovery 34(5):1454--1495

\bibitem[{Dem{\v{s}}ar(2006)}]{demvsar2006statistical}
Dem{\v{s}}ar J (2006) Statistical comparisons of classifiers over multiple data
  sets. Journal of Machine learning research 7(Jan):1--30

\bibitem[{Deng et~al.(2013)Deng, Runger, Tuv, and Vladimir}]{deng2013time}
Deng H, Runger G, Tuv E, Vladimir M (2013) A time series forest for
  classification and feature extraction. Information Sciences 239:142--153

\bibitem[{Drucker et~al.(1997)Drucker, Burges, Kaufman, Smola, and
  Vapnik}]{drucker1997support}
Drucker H, Burges CJ, Kaufman L, Smola AJ, Vapnik V (1997) Support vector
  regression machines. In: Advances in neural information processing systems,
  pp 155--161

\bibitem[{Dua and Graff(2017)}]{Dua:2019}
Dua D, Graff C (2017) {UCI} machine learning repository.
  \urlprefix\url{http://archive.ics.uci.edu/ml}

\bibitem[{Fawaz et~al.(2018)Fawaz, Forestier, Weber, Idoumghar, and
  Muller}]{fawaz2018transfer}
Fawaz HI, Forestier G, Weber J, Idoumghar L, Muller PA (2018) Transfer learning
  for time series classification. In: Proceedings of the 2018 {IEEE}
  International Conference on Big Data (Big Data), pp 1367--1376

\bibitem[{Fawaz et~al.(2019)Fawaz, Forestier, Weber, Idoumghar, and
  Muller}]{fawaz2019deep}
Fawaz HI, Forestier G, Weber J, Idoumghar L, Muller PA (2019) Deep learning for
  time series classification: a review. Data Mining and Knowledge Discovery
  33(4):917--963

\bibitem[{Fawaz et~al.(2020)Fawaz, Lucas, Forestier, Pelletier, Schmidt, Weber,
  Webb, Idoumghar, Muller, and Petitjean}]{fawaz2019inceptiontime}
Fawaz HI, Lucas B, Forestier G, Pelletier C, Schmidt DF, Weber J, Webb GI,
  Idoumghar L, Muller PA, Petitjean F (2020) Inceptiontime: Finding alexnet for
  time series classification. Data Mining and Knowledge Discovery
  34(6):1936--1962

\bibitem[{Friedman(1940)}]{friedman1940comparison}
Friedman M (1940) A comparison of alternative tests of significance for the
  problem of m rankings. The Annals of Mathematical Statistics 11(1):86--92

\bibitem[{Fulcher et~al.(2013)Fulcher, Little, and
  Jones}]{fulcher_little_jones_2013}
Fulcher BD, Little MA, Jones NS (2013) Highly comparative time-series analysis:
  the empirical structure of time series and their methods. Journal of The
  Royal Society Interface 10(83):20130048, \doi{10.1098/rsif.2013.0048},
  \urlprefix\url{https://dx.doi.org/10.1098/rsif.2013.0048}

\bibitem[{Gardner~Jr(1985)}]{gardner1985exponential}
Gardner~Jr ES (1985) {Exponential smoothing: The state of the art}. Journal of
  Forecasting 4(1):1--28

\bibitem[{Goldsmith and Scheipl(2014)}]{goldsmith2014estimator}
Goldsmith J, Scheipl F (2014) Estimator selection and combination in
  scalar-on-function regression. Computational Statistics \& Data Analysis
  70:362--372

\bibitem[{Grabocka et~al.(2014)Grabocka, Schilling, Wistuba, and
  Schmidt-Thieme}]{grabocka2014learning}
Grabocka J, Schilling N, Wistuba M, Schmidt-Thieme L (2014) Learning
  time-series shapelets. In: Proceedings of the 20th ACM SIGKDD International
  Conference on Knowledge Discovery and Data Mining, pp 392--401

\bibitem[{Hyndman(2018)}]{hyndman2018brief}
Hyndman R (2018) A brief history of time series forecasting competitions

\bibitem[{Hyndman et~al.(2008)Hyndman, Koehler, Ord, and
  Snyder}]{hyndman2008forecasting}
Hyndman R, Koehler AB, Ord JK, Snyder RD (2008) Forecasting with Exponential
  Smoothing: The State Space Approach. Springer Science \& Business Media

\bibitem[{Kang et~al.(2017)Kang, Hyndman, and
  Smith-Miles}]{kang_hyndman_smith-miles_2017}
Kang Y, Hyndman RJ, Smith-Miles K (2017) Visualising forecasting algorithm
  performance using time series instance spaces. International Journal of
  Forecasting 33(2):345–358, \doi{10.1016/j.ijforecast.2016.09.004},
  \urlprefix\url{https://dx.doi.org/10.1016/j.ijforecast.2016.09.004}

\bibitem[{Karlen et~al.(2010)Karlen, Turner, Cooke, Dumont, and
  Ansermino}]{karlen2010capnobase}
Karlen W, Turner M, Cooke E, Dumont G, Ansermino JM (2010) Capnobase: Signal
  database and tools to collect, share and annotate respiratory signals. In:
  Annual Meeting of the Society for Technology in Anesthesia (STA), West Palm
  Beach, p~25

\bibitem[{Lin et~al.(2012)Lin, Khade, and Li}]{lin2012rotation}
Lin J, Khade R, Li Y (2012) Rotation-invariant similarity in time series using
  bag-of-patterns representation. Journal of Intelligent Information Systems
  39(2):287--315

\bibitem[{Lines and Bagnall(2015)}]{lines2015time}
Lines J, Bagnall A (2015) Time series classification with ensembles of elastic
  distance measures. Data Mining and Knowledge Discovery 29(3):565--592

\bibitem[{Lines et~al.(2012)Lines, Davis, Hills, and
  Bagnall}]{lines2012shapelet}
Lines J, Davis LM, Hills J, Bagnall A (2012) A shapelet transform for time
  series classification. In: Proceedings of the 18th ACM SIGKDD International
  Conference on Knowledge Discovery and Data Mining, pp 289--297

\bibitem[{Lines et~al.(2016)Lines, Taylor, and Bagnall}]{lines2016hive}
Lines J, Taylor S, Bagnall A (2016) {HIVE-COTE}: The hierarchical vote
  collective of transformation-based ensembles for time series classification.
  In: Proceedings of the 16th {IEEE} International Conference on Data Mining
  (ICDM), pp 1041--1046

\bibitem[{Lubba et~al.(2019)Lubba, Sethi, Knaute, Schultz, Fulcher, and
  Jones}]{lubba_sethi_knaute_schultz_fulcher_jones_2019}
Lubba CH, Sethi SS, Knaute P, Schultz SR, Fulcher BD, Jones NS (2019) catch22:
  Canonical time-series characteristics. Data Mining and Knowledge Discovery
  33(6):1821–1852, \doi{10.1007/s10618-019-00647-x}

\bibitem[{Lucas et~al.(2019)Lucas, Shifaz, Pelletier, O’Neill, Zaidi,
  Goethals, Petitjean, and Webb}]{lucas2019proximity}
Lucas B, Shifaz A, Pelletier C, O’Neill L, Zaidi N, Goethals B, Petitjean F,
  Webb GI (2019) Proximity forest: an effective and scalable distance-based
  classifier for time series. Data Mining and Knowledge Discovery
  33(3):607--635

\bibitem[{Makridakis and Hibon(2000)}]{makridakis2000m3}
Makridakis S, Hibon M (2000) The {M3}-competition: results, conclusions and
  implications. International Journal of Forecasting 16(4):451--476

\bibitem[{Makridakis et~al.(1982)Makridakis, Andersen, Carbone, Fildes, Hibon,
  Lewandowski, Newton, Parzen, and Winkler}]{makridakis1982accuracy}
Makridakis S, Andersen A, Carbone R, Fildes R, Hibon M, Lewandowski R, Newton
  J, Parzen E, Winkler R (1982) The accuracy of extrapolation (time series)
  methods: Results of a forecasting competition. Journal of Forecasting
  1(2):111--153

\bibitem[{Makridakis et~al.(2018)Makridakis, Spiliotis, and
  Assimakopoulos}]{makridakis2018m4}
Makridakis S, Spiliotis E, Assimakopoulos V (2018) The {M4} competition:
  Results, findings, conclusion and way forward. International Journal of
  Forecasting 34(4):802--808

\bibitem[{Makridakis et~al.(2020)Makridakis, Spiliotis, and
  Assimakopoulos}]{makridakis2020m4}
Makridakis S, Spiliotis E, Assimakopoulos V (2020) The {M4} competition:
  100,000 time series and 61 forecasting methods. International Journal of
  Forecasting 36(1):54--74

\bibitem[{Meredith et~al.(2012)Meredith, Clifton, Charlton, Brooks, Pugh, and
  Tarassenko}]{meredith2012photoplethysmographic}
Meredith DJ, Clifton D, Charlton P, Brooks J, Pugh C, Tarassenko L (2012)
  Photoplethysmographic derivation of respiratory rate: a review of relevant
  physiology. Journal of Medical Engineering \& Technology 36(1):1--7

\bibitem[{Moniz and Torgo(2018)}]{moniz2018multi}
Moniz N, Torgo L (2018) Multi-source social feedback of online news feeds.
  arXiv preprint arXiv:180107055

\bibitem[{Montero-Manso et~al.(2020)Montero-Manso, Athanasopoulos, Hyndman, and
  Talagala}]{montero-manso_athanasopoulos_hyndman_talagala_2020}
Montero-Manso P, Athanasopoulos G, Hyndman RJ, Talagala TS (2020) Fforma:
  Feature-based forecast model averaging. International Journal of Forecasting
  36(1):86–92, \doi{10.1016/j.ijforecast.2019.02.011}

\bibitem[{Mueen et~al.(2011)Mueen, Keogh, and Young}]{mueen2011logical}
Mueen A, Keogh E, Young N (2011) Logical-shapelets: an expressive primitive for
  time series classification. In: Proceedings of the 17th {ACM} {SIGKDD}
  International Conference on Knowledge Discovery and Data Mining, pp
  1154--1162

\bibitem[{Nielsen(2016)}]{nielsen2016tree}
Nielsen D (2016) Tree boosting with xgboost-why does xgboost win every machine
  learning competition? Master's thesis, NTNU

\bibitem[{Pedregosa et~al.(2011)Pedregosa, Varoquaux, Gramfort, Michel,
  Thirion, Grisel, Blondel, Prettenhofer, Weiss, Dubourg, Vanderplas, Passos,
  Cournapeau, Brucher, Perrot, and Duchesnay}]{scikit-learn}
Pedregosa F, Varoquaux G, Gramfort A, Michel V, Thirion B, Grisel O, Blondel M,
  Prettenhofer P, Weiss R, Dubourg V, Vanderplas J, Passos A, Cournapeau D,
  Brucher M, Perrot M, Duchesnay E (2011) Scikit-learn: Machine learning in
  {P}ython. Journal of Machine Learning Research 12:2825--2830

\bibitem[{Pelletier et~al.(2019)Pelletier, Webb, and
  Petitjean}]{pelletier2019temporal}
Pelletier C, Webb GI, Petitjean F (2019) Temporal convolutional neural network
  for the classification of satellite image time series. Remote Sensing
  11(5):523

\bibitem[{Pimentel et~al.(2015)Pimentel, Charlton, and
  Clifton}]{pimentel2015probabilistic}
Pimentel MA, Charlton PH, Clifton DA (2015) Probabilistic estimation of
  respiratory rate from wearable sensors. In: Wearable Electronics Sensors,
  Springer, pp 241--262

\bibitem[{Pimentel et~al.(2016)Pimentel, Johnson, Charlton, Birrenkott,
  Watkinson, Tarassenko, and Clifton}]{pimentel2016toward}
Pimentel MA, Johnson AE, Charlton PH, Birrenkott D, Watkinson PJ, Tarassenko L,
  Clifton DA (2016) Toward a robust estimation of respiratory rate from pulse
  oximeters. IEEE Transactions on Biomedical Engineering 64(8):1914--1923

\bibitem[{Rakthanmanon and Keogh(2013)}]{rakthanmanon2013fast}
Rakthanmanon T, Keogh E (2013) Fast shapelets: A scalable algorithm for
  discovering time series shapelets. In: Proceedings of the 2013 SIAM
  International Conference on Data Mining (SDM), SIAM, pp 668--676

\bibitem[{Reiss et~al.(2019)Reiss, Indlekofer, Schmidt, and
  Van~Laerhoven}]{reiss2019deep}
Reiss A, Indlekofer I, Schmidt P, Van~Laerhoven K (2019) Deep {PPG}:
  Large-scale heart rate estimation with convolutional neural networks. Sensors
  19(14):3079

\bibitem[{Reiss et~al.(2017)Reiss, Goldsmith, Shang, and
  Ogden}]{reiss2017methods}
Reiss PT, Goldsmith J, Shang HL, Ogden RT (2017) Methods for scalar-on-function
  regression. International Statistical Review 85(2):228--249

\bibitem[{Salehizadeh et~al.(2016)Salehizadeh, Dao, Bolkhovsky, Cho, Mendelson,
  and Chon}]{salehizadeh2016novel}
Salehizadeh S, Dao D, Bolkhovsky J, Cho C, Mendelson Y, Chon KH (2016) A novel
  time-varying spectral filtering algorithm for reconstruction of motion
  artifact corrupted heart rate signals during intense physical activities
  using a wearable photoplethysmogram sensor. Sensors 16(1):10

\bibitem[{Sammut and Webb(2011)}]{sammut2011encyclopedia}
Sammut C, Webb GI (2011) Encyclopedia of machine learning. Springer Science \&
  Business Media

\bibitem[{Sch{\"a}ck et~al.(2017)Sch{\"a}ck, Muma, and
  Zoubir}]{schack2017computationally}
Sch{\"a}ck T, Muma M, Zoubir AM (2017) Computationally efficient heart rate
  estimation during physical exercise using photoplethysmographic signals. In:
  2017 25th European Signal Processing Conference (EUSIPCO), IEEE, pp
  2478--2481

\bibitem[{Sch{\"a}fer(2015)}]{schafer2015boss}
Sch{\"a}fer P (2015) The {BOSS} is concerned with time series classification in
  the presence of noise. Data Mining and Knowledge Discovery 29(6):1505--1530

\bibitem[{Sch{\"a}fer and Leser(2017{\natexlab{a}})}]{schafer2017fast}
Sch{\"a}fer P, Leser U (2017{\natexlab{a}}) Fast and accurate time series
  classification with weasel. In: Proceedings of the 2017 ACM on Conference on
  Information and Knowledge Management, pp 637--646

\bibitem[{Sch{\"a}fer and Leser(2017{\natexlab{b}})}]{schafer2017multivariate}
Sch{\"a}fer P, Leser U (2017{\natexlab{b}}) Multivariate time series
  classification with {WEASEL}+{MUSE}. arXiv preprint arXiv:171111343

\bibitem[{Segal(2004)}]{segal2004machine}
Segal MR (2004) Machine learning benchmarks and random forest regression. UCSF:
  Center for Bioinformatics and Molecular Biostatistics

\bibitem[{Senin and Malinchik(2013)}]{senin2013sax}
Senin P, Malinchik S (2013) {SAX-VSM}: Interpretable time series classification
  using {SAX} and vector space model. In: 2013 IEEE 13th international
  conference on data mining, IEEE, pp 1175--1180

\bibitem[{Shokoohi-Yekta et~al.(2017)Shokoohi-Yekta, Hu, Jin, Wang, and
  Keogh}]{shokoohi2017generalizing}
Shokoohi-Yekta M, Hu B, Jin H, Wang J, Keogh E (2017) Generalizing dtw to the
  multi-dimensional case requires an adaptive approach. Data Mining and
  Knowledge Discovery 31(1):1--31

\bibitem[{Tan et~al.(2018)Tan, Herrmann, Forestier, Webb, and
  Petitjean}]{tan2018efficient}
Tan CW, Herrmann M, Forestier G, Webb GI, Petitjean F (2018) Efficient search
  of the best warping window for dynamic time warping. In: Proceedings of the
  2018 {SIAM} International Conference on Data Mining (SDM), SIAM, pp 225--233

\bibitem[{Tan et~al.(2020{\natexlab{a}})Tan, Bergmeir, Petitjean, and
  Webb}]{tan2020monash}
Tan CW, Bergmeir C, Petitjean F, Webb GI (2020{\natexlab{a}}) Monash
  {University}, {UEA}, {UCR Time Series Extrinsic Regression Archive}. arXiv
  preprint arXiv:200610996

\bibitem[{Tan et~al.(2020{\natexlab{b}})Tan, Petitjean, and
  Webb}]{tan2020fastee}
Tan CW, Petitjean F, Webb GI (2020{\natexlab{b}}) {FastEE}: Fast ensembles of
  elastic distances for time series classification. Data Mining and Knowledge
  Discovery 34(1):231--272

\bibitem[{Wang et~al.(2017)Wang, Yan, and Oates}]{wang2017time}
Wang Z, Yan W, Oates T (2017) Time series classification from scratch with deep
  neural networks: A strong baseline. In: 2017 International Joint Conference
  on Neural Networks (IJCNN), IEEE, pp 1578--1585

\bibitem[{Ye and Keogh(2009)}]{ye2009time}
Ye L, Keogh E (2009) Time series shapelets: a new primitive for data mining.
  In: Proceedings of the 15th {ACM} {SIGKDD }international conference on
  Knowledge Discovery and Data Mining, pp 947--956

\bibitem[{Yebra et~al.(2018)Yebra, Quan, Ria{\~n}o, Larraondo, van Dijk, and
  Cary}]{yebra2018fuel}
Yebra M, Quan X, Ria{\~n}o D, Larraondo PR, van Dijk AI, Cary GJ (2018) A fuel
  moisture content and flammability monitoring methodology for continental
  australia based on optical remote sensing. Remote Sensing of Environment
  212:260--272

\bibitem[{Zhang(2015)}]{zhang2015photoplethysmography}
Zhang Z (2015) Photoplethysmography-based heart rate monitoring in physical
  activities via joint sparse spectrum reconstruction. IEEE Transactions on
  Biomedical Engineering 62(8):1902--1910

\bibitem[{Zhang et~al.(2014)Zhang, Pi, and Liu}]{zhang2014troika}
Zhang Z, Pi Z, Liu B (2014) Troika: A general framework for heart rate
  monitoring using wrist-type photoplethysmographic signals during intensive
  physical exercise. IEEE Transactions on Biomedical Engineering 62(2):522--531

\end{thebibliography}

\end{document}